%% file: main.tex
\PassOptionsToPackage{unicode}{hyperref}
\PassOptionsToPackage{naturalnames}{hyperref}
\documentclass[10pt,twocolumn,letterpaper]{article}

\usepackage{iccv}
\usepackage{times}
\usepackage{epsfig}
\usepackage[square,comma,sort,numbers]{natbib}
\usepackage{graphicx}
\usepackage{amsmath}
\usepackage{array}
\usepackage{amssymb}
\usepackage{booktabs}
\usepackage{tabularx}
\usepackage{siunitx}
\usepackage{adjustbox}
\usepackage{multirow}
\usepackage{stmaryrd}
\usepackage{pifont}
\usepackage{algorithm}
\usepackage{algpseudocode}
\usepackage{color, colortbl}
\usepackage[dvipsnames]{xcolor}
\usepackage{enumitem,kantlipsum}
\usepackage{enumitem}
\usepackage[pagebackref,breaklinks,colorlinks]{hyperref}
\usepackage[capitalize]{cleveref}
\usepackage{lipsum}

\setlist[enumerate]{label*=\arabic*.}
\newcommand{\cmark}{\ding{51}}%
\newcommand{\xmark}{\ding{55}}%
\definecolor{Gray}{gray}{0.95}

\DeclareMathOperator*{\argmin}{arg\,min}
\newcommand{\ak}[1]{\textcolor{red}{#1}}
\definecolor{Gray}{gray}{0.95}
\definecolor{coolblack}{rgb}{0.0, 0.23, 0.64}
\newcommand{\jnkc}[1]{\textcolor{coolblack}{#1}}

\crefname{section}{Sec.}{Secs.}
\Crefname{section}{Section}{Sections}
\Crefname{table}{Table}{Tables}
\crefname{table}{Tab.}{Tabs.}

\makeatletter
\DeclareRobustCommand{\Arrow}[1][]{%
    \check@mathfonts
    \if\relax\detokenize{#1}\relax
    \settowidth{\dimen@}{$\m@th\rightarrow$}%
    \else
    \setlength{\dimen@}{#1}%
    \fi
    \sbox\z@{\usefont{U}{lasy}{m}{n}\symbol{41}}%
    \begin{picture}(\dimen@,\ht\z@)
    \roundcap
    \put(\dimexpr\dimen@-.7\wd\z@,0){\usebox\z@}
    \put(0,\fontdimen22\textfont2){\line(1,0){\dimen@}}
    \end{picture}%
}
\makeatother

\newcommand{\expectation}{\mathop{\mathbb{E}}}
\newcommand{\PreserveBackslash}[1]{\let\temp=\\#1\let\\=\temp}
\newcolumntype{C}[1]{>{\PreserveBackslash\centering}p{#1}}
\newcolumntype{R}[1]{>{\PreserveBackslash\raggedleft}p{#1}}
\newcolumntype{L}[1]{>{\PreserveBackslash\raggedright}p{#1}}

\iccvfinalcopy
\ificcvfinal\fi
\begin{document}

\title{Domain-Specificity Inducing Transformers for Source-Free Domain Adaptation}
\author{
    Sunandini Sanyal\thanks{Equal Contribution} ~ Ashish Ramayee Asokan\footnotemark[1] ~ Suvaansh Bhambri\footnotemark[1] ~ Akshay Kulkarni \\
    Jogendra Nath Kundu ~ R Venkatesh Babu \\
    Vision and AI Lab, Indian Institute of Science, Bengaluru
}
\maketitle
\ificcvfinal\thispagestyle{empty}\fi

\begin{abstract}
   Conventional Domain Adaptation (DA) methods aim to learn domain-invariant feature representations to improve the target adaptation performance. However, we motivate that \textit{domain-specificity} is equally important since in-domain trained models hold crucial domain-specific properties that are beneficial for adaptation. Hence, we propose to build a framework that supports disentanglement and learning of \textit{domain-specific factors} and \textit{task-specific factors} in a unified model. Motivated by the success of vision transformers in several multi-modal vision problems, we find that queries could be leveraged to extract the domain-specific factors. Hence, we propose a novel \textit{Domain-Specificity inducing Transformer} (DSiT) framework \footnote{Project Page: \url{http://val.cds.iisc.ac.in/DSiT-SFDA/}} for disentangling and learning both domain-specific and task-specific factors. To achieve disentanglement, we propose to construct novel \textit{Domain-Representative Inputs} (DRI) with domain-specific information to train a domain classifier with a novel domain token. We are the first to utilize vision transformers for domain adaptation in a privacy-oriented source-free setting, and our approach achieves state-of-the-art performance on single-source, multi-source, and multi-target benchmarks. 
\end{abstract}

\section{Introduction}
Machine learning models often fail to generalize to unseen domains due to the discrepancy between training (source) and test (target) data distributions (\ie \textit{domain shift}). This results in poor deployment performance and is also critical for applications like autonomous driving \cite{cordts2016cityscapes} or medical imaging \cite{dou2018unsupervised}. Unsupervised domain adaptation (DA) techniques seek to address this challenging scenario by transferring task-specific knowledge from a labeled source domain to an unlabeled target domain. However, DA works \cite{ganin2016domain} require concurrent access to source and target data. Such a constraint is highly impractical since data tends to be proprietary and cannot be easily shared. Hence, we focus on Source-Free DA \cite{kundu2020universal} that operates under a practical setting of sharing only the source model between a \textit{vendor} and a \textit{client}, without any data-sharing.

Conventional Domain Adaptation works \cite{ganin2016domain} aim to learn task-discriminative features that are domain-invariant, \ie indistinguishable \wrt domain shift. Intuitively, if the feature distributions of the source and target domain are similar, then a low error on the source domain translates to the target, shown theoretically by \cite{ben2010theory}. But domain-invariance does not always result in optimal target performance because a specific domain might be far away from the support of a domain-invariant model \cite{dubey2021adaptive}. Further, supervised in-domain trained models (where train and test datasets come from the same domain) usually perform better as they hold useful domain-specific properties. Thus, we motivate the concept of \textit{domain-specificity to improve the target adaptation performance}.

\begin{figure}
\centering
        \includegraphics[width=0.8\linewidth]{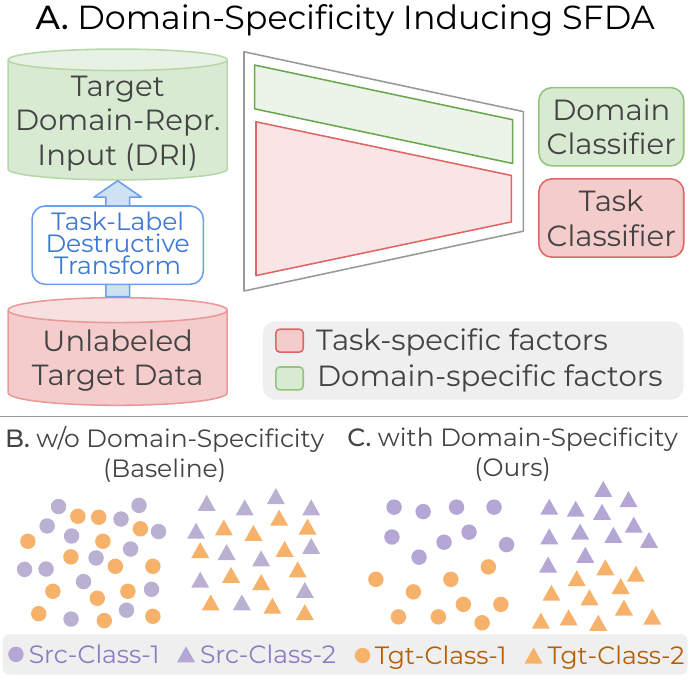}
        \caption{\textbf{A.} We induce domain-specificity by disentangling domain- and task-specific factors within the model. A task-label-destructive transform produces novel Domain-Representative Inputs (DRI) to learn domain-specific factors via domain classification. \textbf{B.} Conventional DA methods preserve domain-invariance, resulting in only task-oriented clusters in the feature space. \textbf{C.} Our proposed disentanglement ensures that different domains are well-clustered.}
        \vspace{-1em}
       \label{fig:teaser}
\end{figure}

In source-free DA, while adapting a model to a new target domain, the major problem is to preserve the task knowledge from the source domain. Prior works address this by aligning the target model feature space with that of the source \cite{SHOT}. However, we argue that it is equally important for a model to learn domain-specific information while preserving task-specific knowledge. Hence, in our work, we seek a solution to an important question, \textit{``How do we develop a framework that enables us to improve domain-specificity while also retaining task-specificity?"}. 

Thus, we seek a framework that supports the disentanglement of task-specific factors and domain-specific factors, thereby providing us better control over them. A well-disentangled framework would allow the learning of both domain-specific and task-specific factors in the model simultaneously, as shown in Fig.\ \ref{fig:teaser}\ak{A}. This not only yields better performance but also ensures that the different domains are well-clustered in the feature space of the model (Fig.\ \ref{fig:teaser}\ak{C}) compared to the baseline approach (Fig.\ \ref{fig:teaser}\ak{B}). However, the key question remains, \textit{``How do we devise a method that encourages the disentanglement and learning of domain-specific and task-specific factors for better adaptation?"}

Recently, vision transformers have demonstrated remarkable performance in several vision tasks \cite{liu2021survey, khan2021transformers}. They contain a multi-head self-attention mechanism that attends to all image patches and provides a global context. Motivated by this fundamental difference of a global context, we explore the possibility of a disentanglement framework with transformers since the domain information is inherently a global, higher-order statistic \cite{chen2020HoMM} that may not be adequately captured in CNNs. Inspired by the multi-modal works \cite{hu2021unit} that utilize queries to extract domain-specific information from a particular modality, we propose to enable the disentanglement of domain-specific and task-specific features through the query weights as part of our novel \textit{Domain-Specificity inducing Transformer} (DSiT) framework.

Concretely, we induce domain-specificity by updating only the query weights via domain classifier training (Fig.\ref{fig:teaser}\textcolor{red}{A}). The remaining weights are updated via task classifier training. To further inculcate the disentanglement, we train the domain-classifier with novel \textit{Domain-Representative Inputs} (DRI) where a task-label-destructive transform removes the task-specific information. We also propose a novel \textit{domain-specificity disentanglement criterion} to evaluate the disentanglement of domain-specific and task-specific factors and demonstrate the disentanglement induced by our proposed framework, DSiT. We outline the contributions of our work as follows:
\begin{itemize}
    
    \item We investigate and provide insights on how domain-specificity can be leveraged to improve DA. To this end, we propose a novel, unified Domain-specificity-inducing Transformer (DSiT) to disentangle and learn task-specific and domain-specific factors.

    \item We utilize query weights to enable the disentanglement in DSiT with a novel training algorithm that well supports our insights. We also introduce novel Domain-Representative Inputs (DRI) to further enhance the disentanglement.

    \item We define a novel domain-specificity disentanglement criterion to determine if the domain-specific and task-specific factors are well-disentangled.
    
    \item We achieve state-of-the-art performance on source-free benchmarks across single-source, multi-source, and multi-target DA while also introducing the first source-free DA benchmarks for transformers. 
\end{itemize}

\section{Related Works}

\noindent
\textbf{Domain Adaptation.}
Prior DA works minimize the domain gap via adversarial distribution matching \cite{hoffman2018cycada, hu2018duplex} or minimize statistical distances \cite{saenko2010adapting, zhang2013domain, saito2018maximum, M3SDA}. However, SFDA works \cite{SHOT, SHOT++, kundu2022subsidiary, kundu2022concurrent, kundu2022balancing} use information maximization and pseudo-labeling to match the target features with the source features. \cite{NRC, li2020model} perform neighborhood clustering and regularization in the target domain. Our work also considers an SFDA setup \cite{kundu2021generalize, kundu2020universal} that focuses on a practical vendor-client setting where vendor and client may use cooperative/same learning strategies, without data sharing.

\vspace{1mm}
\noindent
\textbf{Transformer-based DA.}
Vision transformers are self-attention-based architectures with 
state-of-the-art performance on several vision tasks like object recognition \cite{carion2020end}, semantic segmentation \cite{hoyer2022daformer}, \etc. The application of transformers in DA scenarios is relatively less explored. CDTrans \cite{xu2021cdtrans} proposes cross-attention between source and target image pairs for domain alignment. SSRT \cite{sun2022safe} uses a self-training mechanism while \cite{yang2021tvt} does adversarial training. TransDA \cite{transDA} incorporates a self-attention layer on top of a ResNet backbone for SFDA, and hence is not a pure Vision Transformer (ViT) based solution. Unlike prior SFDA works, we propose a ViT-based solution for the first time, with a focus on enhancing domain-specificity.

\vspace{1mm}
\noindent
\textbf{Domain-specificity for DA.}
Prior works such as \cite{draNet, lee2021unsupervised} focus on style and content disentanglement and aim to achieve domain invariance by removing the style (domain-specific) information from images while preserving only the content information. However, we argue that domain invariance doesn't guarantee optimal performance. We draw motivation from domain generalization works \cite{dubey2021adaptive} that propose constructing domain prototypes from unlabeled target samples to learn domain-specific features. Similarly, DiDA \cite{cao2021dida} proposes feature disentanglement into common and domain-specific features to improve adaptation performance. Furthermore, DMG \cite{DMG} proposed to balance specificity and invariance via learning domain-specific masks and \cite{chang2019domain} proposed to learn domain-specific batch-normalization parameters. Inspired by these, we propose to leverage domain-specificity in transformers to improve the target adaptation performance in a source-free DA setting.

\section{Approach}

\textbf{Problem setup.}
For closed-set DA, we consider a labeled source dataset ${{{\mathcal{D}}_s = \{( x_s, y_s) : x_s \!\in\! {\mathcal{X}}, y_s \!\in\! {\mathcal{C}}_g\}}}$ where ${\mathcal{X}}$ denotes the input space and $\mathcal{C}_g$ denotes the goal task label set. We denote the unlabeled target dataset by ${{\mathcal{D}}_t = \{ x_t : x_t \!\in \!{\mathcal{X}}\}}$. The task is to predict the label for each target sample $x_t$ from the label set $\mathcal{C}_g$. Following \cite{xu2021cdtrans}, we use ViT-B \cite{dosovitskiy2020image} as the backbone feature extractor $h\!:\!\mathcal{X}\!\to\! \{\mathcal{Z}_c, \mathcal{Z}_1, \mathcal{Z}_2, ... \mathcal{Z}_{N_P}\}$ where $\mathcal{Z}_c$ represents the class token feature-space and $\mathcal{Z}_1, \mathcal{Z}_2, ...\mathcal{Z}_{N_P}$ are patch token feature-spaces ($N_P$ is the number of patches). A goal task classifier trained on the class token is denoted as $f_g: \mathcal{Z}_c\to \mathcal{C}_g$. In the source-free vendor-client setup \cite{kundu2020universal}, we operate under the practical constraint where data sharing between vendor and client is prohibited. The vendor trains a model on the source domain and shares only the model with the client. The client uses the vendor-side model to adapt to the target domain. Unlike usual DA works \cite{ganin2016domain} that advocate domain-invariant learning, we investigate how domain-specificity can be leveraged for DA. First, we discuss why domain-specificity is useful through the following insight.

\vspace{1mm}
\noindent
\textbf{Insight 1.\ (Domain-specificity leads to improved DA)}
\noindent
\textit{
In the source-free DA setting, the aim is to achieve good target accuracy for the client-side model. Since an in-domain trained model better represents domain-specific factors as compared to a domain-invariant model, inculcating domain specificity on the client-side plays a major role in improving the target adaptation performance.}

\vspace{1.0mm}
\noindent
\textbf{Remarks.} 
In a vendor-client setting, a client wishes to improve the goal task performance of the model on the target data. Conventional DA methods \cite{ganin2016domain} often devise a strategy to learn domain-invariant features that contain only the task knowledge, which could be used to generalize across multiple domains. However, the optimal task-specific features from a domain-specific model might be far away from the task-specific features of a domain-invariant model \cite{dubey2021adaptive}. Hence, such approaches are unsuitable for source-free DA, where the goal of improved target performance can be better achieved if we inculcate in-domain knowledge to build a domain-specific model. Therefore, we note that it is equally important for a client-side model to learn ``domain-specific factors" containing the crucial in-domain characteristic knowledge along with the ``task-specific factors" that hold information on the goal task. 

As Insight \ak{1} motivates domain-specificity, a natural question arises - ``\textit{How can we enable and control domain-specificity while also learning the task-specific factors?}"

\vspace{4mm}
\noindent
\textbf{Insight 2. (Disentanglement of domain-specific and task-specific factors to control domain specificity)}
\textit{
Disentanglement of domain-specific and task-specific factors provides a way to learn the two orthogonal factors together within the same model, enabling better control over them.}

\noindent
\textbf{Remarks.}
Since source and target domains can be far apart, \ie high variance in domain-specific factors, it is desirable for a domain-specific model to learn a \textit{disentangled} feature space that can adequately capture task-specific as well as domain-specific factors. Prior SFDA works either aim to align the target domain features towards the source domain \cite{SHOT} or update all parameters, risking the loss of source-side task-specific knowledge \cite{NRC}. Conversely, our proposed disentanglement allows simultaneous and separate learning of the domain-specific and task-specific factors. Moreover, these domain-specific factors could be leveraged to orient the task-specific factors \cite{dubey2021adaptive} for a new target domain. 

Next, we seek an answer to the following question, \textit{``How do we devise a unified architecture to enable disentanglement of domain-specific and task-specific factors?"}

\subsection{Exploring the potential of transformers towards domain specificity}
\label{subsec:approach_analysis}
We explore the various possibilities of disentangling domain-specific and task-specific factors in the self-attention module of a transformer. Prior vision transformer works have demonstrated superior feature alignment capabilities within the attention module of a transformer, even across multi-modal scenarios such as text-to-vision \cite{tsai2019multimodal, hu2021unit}. Moreover, transformers are inherently more robust to noise \cite{xu2021cdtrans} and capture long-range dependencies via self-attention across different patches compared to the localized context that is captured in CNNs. Hence, we propose to adopt transformer architectures for inducing domain-specificity and are the first to use them in a source-free DA setting. 

\vspace{0.5mm}
\noindent
\textbf{a) Query for domain-specificity.}
Vaswani et al.\ \cite{vaswani2017attention} introduced self-attention as a compatibility function between a Query and a corresponding Key. The final output is computed as a weighted sum of the Value where the weights are assigned as per the self-attention. Prior multi-modal works  \cite{li2021trear,hu2021unit} use modality-specific queries (e.g. from text or image modality) as input to the cross-attention. While we draw inspiration from these works, our novel approach explicitly enforces domain-specificity in the ``self-attention query" (\ie inside every transformer layer), different from the cross-attention approach of multi-modal works. This helps to better control the task-specific factors, where a domain-specific Query learns to acquire the domain-specific information from the input Value, making the overall output domain-specific in nature. While there could be more sophisticated ways to inculcate domain-specificity, such as hypernetworks \cite{ha2016hypernetworks} or auxiliary models \cite{huang2018auggan}, we propose this simple strategy that validates our insights. 

\begin{figure}[t]
\centering
    \includegraphics[width=0.9\linewidth]{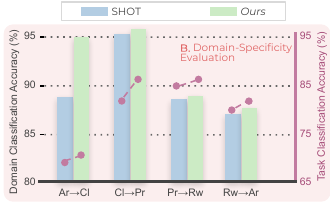}
    \caption{We evaluate the domain-specificity of our adapted model \wrt\ SHOT on Office-Home. Task acc.\ is goal task accuracy and Domain acc.\ is the accuracy of the binary source-target domain classifier (as explained in Sec.\ \ref{subsec:approach_analysis}\textcolor{red}{b}). Our higher task accuracy (in pink) indicates better task-specificity and better disentanglement.}
    \label{fig:a-dist}
    \vspace{-3mm}
\end{figure}

\begin{figure*}[t]
\centering
     \includegraphics[width=1.0\linewidth]{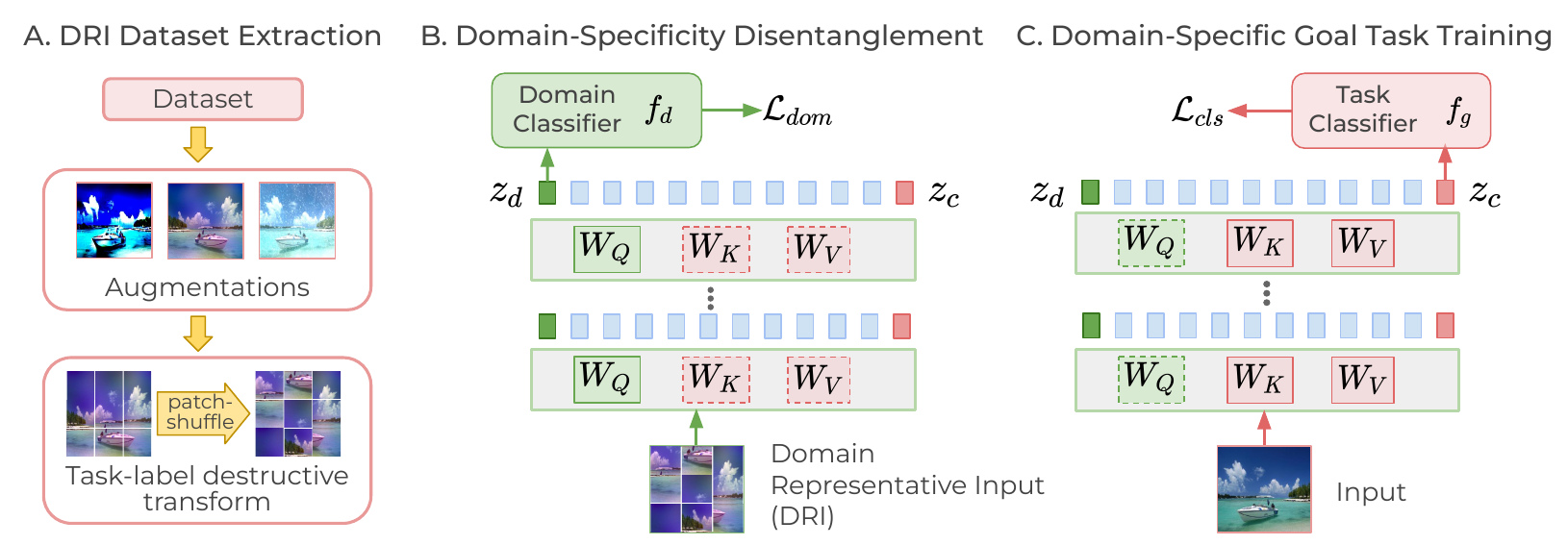}
        \caption{\textbf{DSiT Training:} 
        \textit{\textbf{A.\ DRI dataset extraction:}} DRI are obtained via patch-shuffling of augmented inputs.
        \textit{\textbf{B.\ Domain-Specificity Disentanglement:}} Domain classifier $f_d$ is trained with domain classification loss $\mathcal{L}_{dom}$ using DRI, updating only query weights $W_Q$ via a domain token $z_d$, while other weights including $W_K$ and $W_V$ are frozen (dotted boundary). \textit{\textbf{C.\ Domain-Specific Goal Task Training:}} The task classifier is trained with task classification loss $\mathcal{L}_{cls}$ updating all weights except $W_Q$.
        }
        \vspace{-1em}
       \label{fig:fig_}
\end{figure*}

\vspace{0.5mm}
\noindent
\textbf{b) Evaluating domain-specificity and task-specificity.}
To analyze the effectiveness of our proposed strategy, we examine the domain-specificity of the ViT class token by training a linear domain classifier with the source and target domain samples. The accuracy is computed on four settings Ar$\rightarrow$Cl, Cl$\rightarrow$Pr, Pr$\rightarrow$Rw, and Rw$\rightarrow$Ar on the Office-Home dataset. From Fig. \ref{fig:a-dist}, we observe that our approach achieves a higher domain classification accuracy across all four settings. Note that we access both source and target domain data only for this analysis. Intuitively, we examine the domain-specificity inculcated in the class tokens as they are used by the task classifier. We also report the task performance to highlight the task-specific knowledge of the model in Fig.\ \ref{fig:a-dist}. We observe that our approach achieves higher goal task accuracy as well. This implies that our proposed approach captures the domain-specific and task-specific factors well, leading to better adaptation performance.

\subsection{Training algorithm}
\label{subsec:train_algo}
We propose \textbf{D}omain-\textbf{S}pecificity  \textbf{i}nducing \textbf{T}ransformers for Source-Free DA (\textbf{DSiT}). In the following sections, we describe the proposed approach to train {DSiT}.

\vspace{1mm}
\noindent
\textbf{3.2.1 Vendor-side source training}
\label{subsec:vendorside}

\vspace{1mm}
\noindent
We train {DSiT} in two steps. First, we perform domain-specificity disentanglement via domain classifier training. The second step involves goal task classifier training. Next, we delve into the details of each step.

\vspace{1mm}
\noindent
\textbf{a) Domain-specificity disentanglement} (Fig.\ \ref{fig:fig_}\textcolor{red}{B})
We first induce domain-specificity by training a domain classifier $f_d$ with a novel domain token $z_d \in \mathcal{Z}_d$ from the backbone $h$. For this, the vendor first prepares augmented datasets $\mathcal{D}_s^{(i)} = \{ (x_s^{[i]}, y_s, y_d) \;\forall\; (x_s, y_s) \in \mathcal{D}_s \} \;\forall  \; i \in [N_a]$ by augmenting each source sample $x_s$ ($N_a$ is the number of augmentations). Here, the augmentation $\mathcal{A}_i\!:\! \mathcal{X} \!\to\! \mathcal{X}$ is applied to get $x_s^{[i]} \!=\! \mathcal{A}_i(x_s)$. Each input is assigned a domain label $y_d=i$ where $i$ denotes the augmentation label. We use five label-preserving augmentations that simulate novel domains \cite{kundu2021generalize} (see Suppl. for more details). 

\begin{figure}[t]
\centering
    \includegraphics[width=1.0\linewidth]{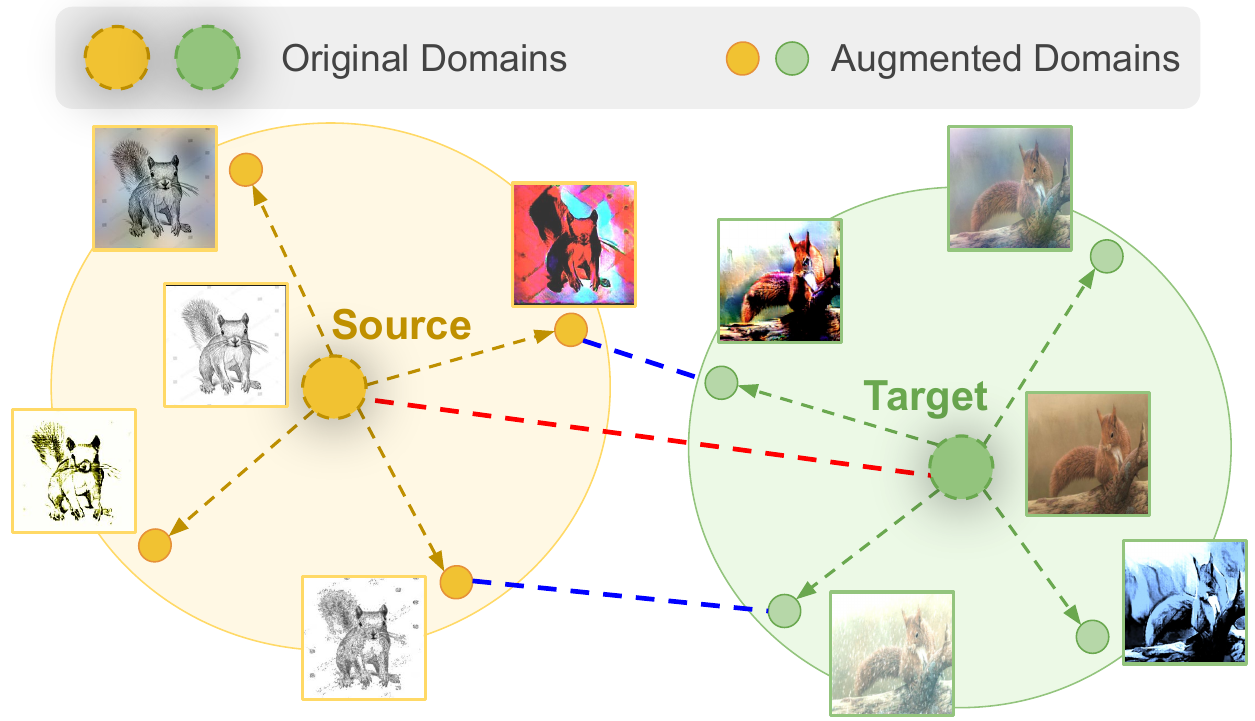}
    \caption{
        Novel augmented domains can be simulated where the domain gap between an augmented source and an augmented target (blue lines) can be less than the original domain gap (red line). With a domain classifier for augmented domains, we induce better domain-specificity as multiple domains with lower domain gaps are well-separated.  
        }
       \label{fig:fig_augs}
       \vspace{-1mm}
\end{figure}

We motivate the use of augmentations with two supports.
First, as we operate in a source-free setting, access to multiple domains cannot be assumed, and these augmented domains can be used to inculcate domain-specificity. Further, this domain-specificity inculcation can be shared between vendor and client by sharing only the augmentation information. Second, as shown in Fig.\ \ref{fig:fig_augs}, an augmented source domain may be closer to another augmented target domain (blue lines) \wrt the original source-target (red line). We also confirm this numerically (reported in Suppl). Thus, we have access to more diverse domains (with lower domain gaps) which need to be well-separated by the domain classifier $f_d$, allowing for better domain-specificity.

While we could train $f_d$ with only augmented samples to inculcate domain-specificity, we also need to disentangle the task-specific information as discussed under Insight \ak{2}.
Thus, we next describe a novel input representation that better represents domain-specific factors by separating the task-specific factors, \ie enabling better disentanglement.

\vspace{1mm}
\noindent
\textbf{{Domain-Representative Input (DRI)}} is constructed at the input level to preserve only domain-specific information by applying a task-label-destructive transformation of patch-shuffling \cite{mitsuzumi2021generalized} as shown in Fig.\ \ref{fig:fig_}\textcolor{red}{A}. Intuitively, domain information is a higher-order statistic \cite{chen2020HoMM} that is preserved after patch-shuffling while class information is lost. The shuffling of patches is extremely vital for amplifying the input domain-specific factors for the domain-specificity training step. Hence, DRIs are not just used as an augmentation for the goal task training, as shown in prior transformer works \cite{qin2021understanding}, but are a means to inculcate domain specificity. We empirically demonstrate in Sec.\ \ref{subsec:expts_analysis}\textcolor{red}{b} that DRI 
guides the goal task performance better than only augmented inputs, because of improved disentanglement. We also provide additional baselines in Table \ref{tab:dri_abl} to validate the unsuitability of DRI as a general augmentation for the goal task training.

\noindent
As per our analysis in Sec.\ \ref{subsec:approach_analysis}, we train only the transformer query weights, to exclusively hold domain-specific knowledge, with the domain classification loss,
\begin{equation}
    \min_{\theta_Q, \theta_{f_d}} \expectation_{(x, y_d) \in  \cup_{i} \mathcal{D}_s^{(i)}} [\mathcal{L}_{dom}] \text{ where } \mathcal{L}_{dom} = \mathcal{L}_{ce}(f_d(z_d), y_d)
    \label{eqn:src_dom_loss}
\end{equation}
\noindent
where $z_d$ is the domain-token output from backbone $h$ and $\theta_Q \!=\! \cup_{j} W_{Q_{j}}$; $j$ is an index over the backbone layers. Note that DRI is used for $\mathcal{L}_{dom}$. For simplicity, we re-use $\mathcal{D}_s^{(i)}$ to include the task-label-destructive transform (Fig.\ \ref{fig:fig_}\textcolor{red}{A}).

\vspace{1mm}
\noindent
{\textbf{b) Domain-specific goal task training}} (Fig.\ \ref{fig:fig_}\textcolor{red}{C}) After one round of domain classifier training, we train DSiT for the goal task. Here, we update key and value weights $W_K$ and $W_V$ along with all other parameters (except query weights $W_Q$). Hence, the frozen queries are disentangled and hold crucial domain-specific information. The vendor trains a source model consisting of the backbone $h$ and task classifier $f_g$ using the source dataset $\mathcal{D}_s$ with the task-classification loss as follows,
\begin{equation}
    \min_{\theta_h\setminus\theta_Q, \theta_{f_g}} \expectation_{(x_s, y_s) \in  \mathcal{D}_s} [\mathcal{L}_{cls}] 
    \text{ where } \mathcal{L}_{cls} = \mathcal{L}_{ce}(f_g(z_c), y_c)
    \label{eqn:src_clsf_loss}
\end{equation}

\noindent
where $z_c$ is the class-token output from backbone $h$, and $\theta_h\setminus\theta_Q$ are the parameters of the backbone $h$ excluding all $W_Q$ weights while $\theta_{f_g}$ are the parameters of classifier $f_g$. The two steps of domain-specificity disentanglement (Eq.\ \ref{eqn:src_dom_loss}) and goal task training (Eq.\ \ref{eqn:src_clsf_loss}) are performed in tandem, one after the other. See Suppl.\ for details.

\vspace{4mm}
\vspace{1mm}

\noindent
\textbf{3.2.2 Client-side target adaptation}
\vspace{1mm}

\noindent
The vendor shares the source-trained, disentangled DSiT model with the client for target adaptation. The client also follows the same process of domain-specificity disentanglement as the vendor. To disentangle the target-domain-specific factors and task-specific factors for improved target adaptation (as per Insight \textcolor{red}{1}), the client applies the same augmentations to the target data to generate DRIs for domain classifier training with the domain classification loss $\mathcal{L}_{dom}$ (as described in Sec.\ \ref{subsec:vendorside}\textcolor{red}{.1a} and Fig.\ \ref{fig:fig_}\textcolor{red}{B}). For goal task training, the client uses the standard information maximization and diversity losses \cite{SHOT, GSFDA, NRC}. The overall client-side objective is as follows:
\begin{equation}
    \min_{\theta_h\setminus\theta_Q, \theta_{f_g}} \expectation_{\mathcal{D}_t}  [\mathcal{L}_{im} + \mathcal{L}_{div}] + \min_{\theta_Q, \theta_{f_d}}\expectation_{\cup_{i} \mathcal{D}_t^{(i)}} [\mathcal{L}_{dom}]
\end{equation}

\noindent
We detail $\mathcal{L}_{im}$ and $\mathcal{L}_{div}$ in the Suppl.\ as these are commonly used and not our main contributions. Note that only original target data $\mathcal{D}_t$ is used for $\mathcal{L}_{im}$ and $\mathcal{L}_{div}$. As in Eq.\ \ref{eqn:src_dom_loss}, DRI are used for $\mathcal{L}_{dom}$ here. For simplicity, we re-use $\mathcal{D}_t^{(i)}$ to include the task-label-destructive transform (Fig.\ \ref{fig:fig_}\textcolor{red}{A}).

\input{tables/ssda_oh}

\subsection{Domain-specificity disentanglement criterion}
To evaluate the disentanglement of domain-specific and task-specific factors, we propose a criterion considering the following three feature-space cosine similarity metrics. Let $\gamma_{cls}$ be the intra-class, inter-domain similarity, $\gamma_{dom}$ be the intra-domain, inter-class similarity, and $\gamma_{all}$ be the inter-class, inter-domain similarity. These are computed as the similarity between pairs of features averaged over the possible input pairs for each input. For example, possible input pairs for $\gamma_{cls}$ have to come from the same class but different domains. See Suppl.\ for more details. Based on these, we define a criterion to determine if the domain-specific and task-specific factors of the model are well-disentangled.

\vspace{1mm}
\noindent
\textbf{Definition 1. (Domain-Specificity Disentanglement Criterion)} \textit{The domain-specific and task-specific factors of a model are well-disentangled if}
\begin{equation}
    \gamma_{cls}, \gamma_{dom} > \gamma_{all} \text{ and } \vert \gamma_{cls}-\gamma_{dom}\vert < \tau
\end{equation}

\vspace{1mm}
\noindent
\textbf{Remarks.}
Here, $\tau$ is a threshold. In other words, $\gamma_{cls}$ and $\gamma_{dom}$ should be higher than $\gamma_{all}$ and the absolute difference between $\gamma_{cls}$ and $\gamma_{dom}$ should be low. Based on the definitions, $\gamma_{cls}$ indicates task-specificity since the used samples belong to the same class but different domains. Similarly, $\gamma_{dom}$ represents the domain-specificity. $\gamma_{all}$ is the similarity between samples from different classes and different domains, which should be lower than $\gamma_{dom}$ and $\gamma_{cls}$ with any domain-specific and task-specific knowledge, respectively. Along with the first condition, if the difference between task-specificity and domain-specificity is low, \ie both are equally present, the two will be well-disentangled in the model. {In Sec.\ \ref{subsec:expts_analysis}\textcolor{red}{c}, we empirically verify that our proposed approach satisfies Definition \ak{1}.}

\section{Experiments}
\noindent
We evaluate the effectiveness of our proposed approach on the standard DA benchmarks across several settings. 

\vspace{0.5mm}
\noindent
\textbf{Datasets.}
We demonstrate the efficacy of our work on the following three standard object recognition DA benchmarks. \textbf{Office-31} \cite{saenko2010adapting} is a benchmark consisting of three domains under office environments - Amazon (\textbf{A}), DSLR (\textbf{D}), and Webcam (\textbf{W}) with 31 classes each. \textbf{Office-Home } \cite{venkateswara2017deep} consists of images of everyday objects divided into four domains $-$ Artistic (\textbf{Ar}), ClipArt (\textbf{Cl}), Product (\textbf{Pr}), and Real-world (\textbf{Rw}), each with 65 classes. \textbf{VisDA } \cite{visda} is a large-scale dataset for adapting algorithms from synthetic domains to real domains. There are 152,397 synthetic images in the source domain and 55,388 real-world images in the target domain. \textbf{DomainNet } \cite{M3SDA} is the most challenging benchmark with six domains, each with 345 classes: ClipArt (\textbf{C}), Real (\textbf{R}), Infograph (\textbf{I}), Painting (\textbf{P}), Sketch (\textbf{S}), and Quickdraw (\textbf{Q}).

\input{tables/ssda_o31}
\input{tables/ssda_dn}

\vspace{1.0mm}
\noindent
\textbf{Implementation details.}
We follow the experimental settings of CDTrans \cite{xu2021cdtrans} and use DeiT-Base \cite{touvron2021training} as the backbone ViT for fair comparisons. The DeiT-Base architecture consists of 12 layers, where each layer consists of 12 self-attention heads collectively termed as multi-head self-attention. We use an input size of $224 \times 224$ and a patch size of $16 \times 16$ for all our experiments, resulting in $14 \times 14$ patches (196 total) per input. DeiT-B contains an additional distillation token, however, the rest of the architecture is the same as a ViT-B backbone. For optimizing the training objectives, we use Stochastic Gradient Descent (SGD) with momentum $0.9$, and a weight decay ratio of $1 \times 10^{-4}$. Refer to the Suppl.\ for the complete implementation details.

\subsection{Comparisons with prior works}
\noindent
We compare with SOTA methods on single-source, and multi-source DA in Table \ref{tab:ssda_oh}, \ref{tab:ssda_o31}, \ref{tab:ssda_dn} and \ref{tab:msda_oh_dn} and multi-target DA (see Suppl. for details).

\input{tables/msda_oh}

\vspace{1mm}
\noindent
\textbf{a) Single-source domain adaptation (SSDA).}
Table \ref{tab:ssda_oh} \ref{tab:ssda_o31}, and \ref{tab:ssda_dn} present comparisons between our proposed approach DSiT and prior SSDA works. For the Office-Home benchmark, our approach achieves state-of-the-art performance among source-free works (Table \ref{tab:ssda_oh}). DSiT outperforms the source-free prior works SHOT-B* by 2.4\%, Mixup by 2\%, and DIPE by 2.3\% on Office-Home and yields comparable performance to the non-source-free transformer-based prior work CDTrans. Similarly, on the Office-31 benchmark (Table \ref{tab:ssda_o31}), our approach improves over the source-free works SHOT-B* by 1.6\%, Mixup by 1.3\% and DIPE by 2.5\% and gives competitive results compared to non-source-free works. On the more challenging VisDA dataset (Table \ref{tab:ssda_o31}), our approach outperforms source-free SHOT-B* baseline by 1.7\%. DSiT also outperforms all the existing source-free methods by a significant margin on the most challenging benchmark DomainNet (Table \ref{tab:ssda_dn}), and also achieves a 5.1\% improvement over CDTrans, despite the latter being non-source-free. Refer to Suppl.\ for the complete table.

\vspace{1mm}
\noindent
\textbf{b) Multi-source domain adaptation (MSDA).} In Table \ref{tab:msda_oh_dn}, we compare our approach with source-free and non-source-free prior works for multi-source DA. We improve over the source-free baseline SHOT-B* by 1\% on Office-Home. Despite the source-free constraint that we address with our method, we outperform even non-source-free works.

\noindent
Overall, these benchmark results highlight the efficacy of the proposed disentanglement of domain-specific and task-specific factors (Insight \ak{2}) across three diverse settings of single-source, multi-source, and multi-target DA.

\subsection{Analysis}
\label{subsec:expts_analysis}
In Table \ref{tab:ablation}, we provide an ablation study for various steps in the training of our approach. 

\vspace{1mm}
\noindent
\textbf{a) Effect of inculcating domain-specificity.}
In Table \ref{tab:ablation}, we compare the effect of domain-specificity training (without DRI) on both vendor-side and client-side training. On vendor-side (\#1 vs.\ \#2), we observe improvements of 0.8\% while the improvements on client-side (\#4 vs.\ \#5) are 1.0\%. This is in line with our Insight \ak{1} that domain-specificity supports target adaptation performance (even without DRI).

\vspace{1mm}
\noindent
\textbf{b) Effect of DRI.} As discussed in Sec.\ \ref{subsec:train_algo}, domain-representative inputs (DRI) reduce the impact of task-specific information while preserving and aiding in the learning of domain-specific information. This further amplifies the impact of domain-specificity on the goal task, which results in improvements of 1\% on the vendor-side (\#2 vs.\ \#3) and 1.1\% on the client-side (\#5 vs.\ \#6).

\input{tables/dri_ablation}
\input{tables/dsit_ablation}

\vspace{1mm}
\noindent
\textbf{c) Empirical analysis of Definition 1.}
As per the discussion under Definition \ak{1}, we report the three similarity metrics for the baseline SHOT and our DSiT in Table \ref{tab:defn1_analysis}. The condition of $\gamma_{dom}, \gamma_{cls} > \gamma_{all}$ is satisfied by both models (since $\gamma_{all}$ denotes similarity between any class and any domain samples which would be lower in most cases). $\gamma_{dom}$ and $\gamma_{cls}$ for DSiT are closer in magnitude than SHOT, indicating its better disentanglement. This is further strengthened by the improved DA performance of our proposed method DSiT compared to the source-free baseline SHOT.

\vspace{1mm}
\noindent
\textbf{d) Performance of DRI as a general augmentation.}
To preserve and enhance domain-specificity, we utilize DRIs only for the domain-classifier training (Eq. \ref{eqn:src_dom_loss}), which helps to improve the target adaptation performance (as per Insight \textcolor{red}{1}). However, DRIs are unsuitable for task-classifier training as the task label information is destroyed by patch-shuffling. In Table \ref{tab:dri_abl}, we use DRI-augmented inputs for task classifier training with SHOT-B \cite{SHOT} and CDTrans \cite{xu2021cdtrans} baselines and observe significant drops of 1.7\% and 4.7\% on Office-Home (4
settings) with respect to the corresponding baselines, respectively. This drop is expected and validates our insights that DRI is not suitable as a general augmentation for DA, but is extremely crucial for domain-specificity training (+1\% improvement using DRI in Table \ref{tab:ablation}).

\input{tables/dom_spec_crit}

\section{Conclusion}
In this work, we study the concept of domain-specificity in source-free DA. We provide insights to analyze how domain-specificity could be leveraged to improve target adaptation performance. We, therefore, propose a novel Domain-specificity-inducing Transformer (DSiT) where we leverage the queries of a vision transformer to induce domain-specificity and train the unified model to enable a disentanglement of task- and domain-specificity. Based on our insights, we also introduce a novel Domain-Specificity Disentanglement criterion to determine if the task-specific and domain-specific factors are well disentangled. The proposed approach outperforms several state-of-the-art benchmarks for single-source, multi-source, and multi-target DA.

\noindent
\textbf{Acknowledgements.} This work was supported by Kotak IISc AI-ML Centre (KIAC).

\appendix
\twocolumn[
    \begin{@twocolumnfalse}
    \begin{center}
        \textbf{\Large Supplementary Material}
        \vspace{1cm}
    \end{center}
    \end{@twocolumnfalse}
]

\begin{figure*}
\centering
      \includegraphics[width=0.85\linewidth]{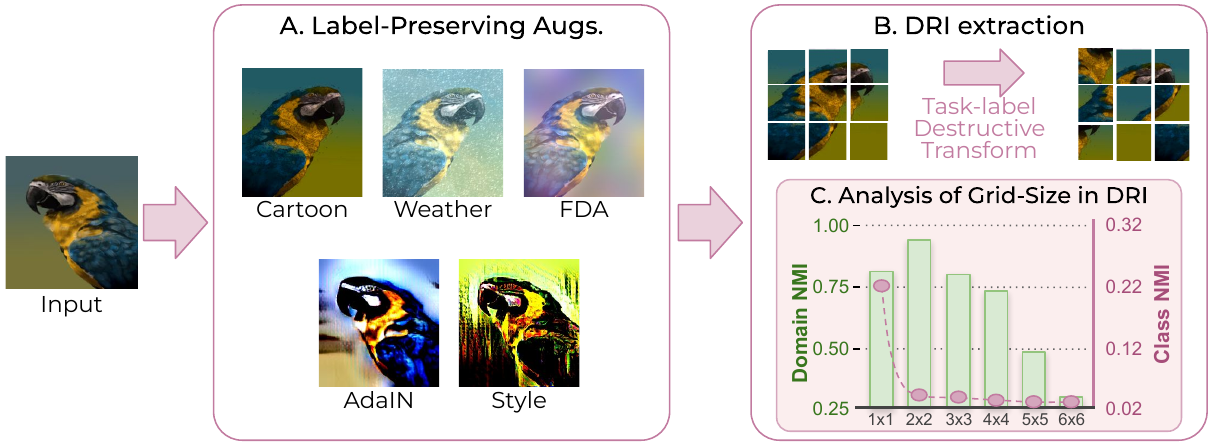}
         \caption{\textbf{A.} Label-preserving augmentations are first applied to the input to simulate novel domains. \textbf{B.} Then, the task-destructive transform of patch-shuffling is used to obtain the DRI image. \textbf{C.} We analyze the Domain-NMI and Class-NMI for different grid-sizes used in patch-shuffling. An example of $3\times 3$ shuffling is shown in B.}
        \label{fig:augs_fig}
\end{figure*}

\noindent
In this supplementary material, we provide more details on the training algorithm, experimental settings, additional comparisons, and analysis experiments. We have released our code on our project page: \url{https://val.cds.iisc.ac.in/DSiT-SFDA/}. The remainder of the supplementary material is structured as shown below:

\begin{itemize}
\vspace{-1mm}
    \item Sec. \ref{sec:shot_loss}: Approach (Algorithm \ref{algo:overall})
    \vspace{-2mm}
    \item Sec. \ref{sec:impl}: Implementation Details
    \vspace{-2mm}
    \begin{itemize}
        \item Sec. \ref{subsec:augs}: Domain Augmentations (Fig. \ref{fig:augs_fig}\textcolor{red}{A})
        
        \item Sec. \ref{subsec:dri_ext}: DRI Dataset Extraction (Fig. \ref{fig:augs_fig}\textcolor{red}{B})
        \item Sec. \ref{sub:dom_spec_crit} Domain-specificity criterion
        \item Sec. \ref{sub:exp_set}: Experimental Settings
    \end{itemize}
    \vspace{-3mm}
    \item Sec. \ref{sub:addn_results} Additional experimental results
    \vspace{-2mm}
    \begin{itemize}
        \item Sec. \ref{sub:ext_comp} Extended comparisons (Table \ref{tab:ssda_dnet}, \ref{tab:mtda_oh})
        \item Sec. \ref{sub:vend_perf} Vendor-side DSiT results (Table \ref{tab:vendor-side})
        \item Sec. \ref{sub:adap_set} Model adaptation setting (Table \ref{tab:so_comp})
        \item Sec. \ref{sub:backbones} Perf. on different backbones (Table \ref{tab:vit-s})
        \item Sec. \ref{sub:augs_analysis} Analysis for augmentations (Table \ref{tab:a-dist augs})
        \item Sec. \ref{sub:sens}: Sensitivity analysis of DSiT (Table \ref{tab:sensitivity_analysis})
        \item Sec. \ref{sub:train_time} Training time comparisons (Table \ref{tab:training_time})
        \item Sec. \ref{sub:stat_sig}: Statistical significance (Table \ref{tab:mean_std})
        \item Sec. \ref{sub:abl_ta}: Target adaptation losses (Table \ref{tab:ablation_loss})
        \item Sec. \ref{sub:dri_size}: Effect of DRI grid-size (Fig. \ref{fig:dri_size})
    \end{itemize}
\end{itemize}

\input{suppl_tables/notations}

\section{Approach}
\label{sec:shot_loss}
Table \ref{sup:tab:notations} shows a complete list of the notations used in the paper. We summarize our full approach in Algorithm \ref{algo:overall} and describe the details of the approach in this section.

\vspace{1mm}
\noindent
\textbf{(a) Target adaptation losses.}
For the client-side target adaptation, we use the Information Maximization loss formulation from SHOT \cite{SHOT}, which consists of two terms: entropy loss $\mathcal{L}_{im}$ and diversity loss $\mathcal{L}_{div}$. The entropy loss $\mathcal{L}_{im}$ ensures that the confidence of the model towards a label is high. The diversity loss $\mathcal{L}_{div}$ ensures that the model's predictions are well-balanced across all classes and prevents the model from producing degenerate solutions. We define the two terms as follows:
\begin{equation}
\label{sup:eqn:loss_im}
    \mathcal{L}_{im} = -\expectation_{x_t \in \mathcal{X}} \sum_{k=1}^{K} \delta_{k}(f_g(z_c)) \log \delta_{k}(f_g(z_c))
\end{equation}
\begin{equation}
\label{sup:eqn:loss_div}
    \mathcal{L}_{div} = \sum_{k=1}^{K} \hat{p_k} \log \hat{p_k}
     = KL (\hat{p}, \frac{1}{K} 1_K) - \log K
\end{equation}
\noindent
where $\delta_{k}(a) = \frac{\exp(a_k)}{\sum_i \exp(a_i)}$ represents the $k^{\text{th}}$ element in the softmax output of $a \in \mathbb{R}^K$, and $z_c$ is the class-token from $h$ for an input $x_t$. We optimize all parameters of the transformer backbone $h$, except the query weights $\theta_Q$ as follows,
\begin{equation}
    \min_{\theta_h\setminus\theta_Q, f_g} \expectation_{\mathcal{D}_t}  [\mathcal{L}_{im} + \mathcal{L}_{div}] 
\end{equation}

\noindent
We also utilize the clustering method of SHOT \cite{SHOT} for self-supervised pseudo-labeling. First, we obtain the class centroids in the target domain via weighted k-means clustering,
\begin{equation}
    c_k = \frac{\sum_{x_t \in \mathcal{X}} \delta_k(f_g(z_c)) z_c}{\sum_{x_t \in \mathcal{X}} \delta_k (f_g(z_c))}
\end{equation}
\noindent
The centroid characterizes the labels for the samples. In order to obtain a pseudo-label, we choose the closest centroid based on the cosine distance as follows,
\begin{equation}
    \hat{y_c} = \argmin_{k} D_c (z_c, c_{k})
\end{equation}
where $D_c$ denotes the cosine-distance in the class-token feature space $\mathcal{Z}_c$ between the centroid and the input sample features $z_c$. As the model keeps training, the centroids are updated after every few iterations, and pseudo-labels are assigned according to the new centroids.

\vspace{1mm}
\noindent
\textbf{(b) Preliminaries on Transformers.}
Recently, Vision Transformers (ViT) have been shown to improve significantly on several vision tasks \cite{dosovitskiy2020image}. Self-attention is one of the most important components in the transformer architecture. A ViT takes an image as input $x \in \mathcal{X}=\mathbb{R}^{H \times W \times C}$ in the form of patches of size $(P,P)$, where $H, W$ is the image size and $C$ is the number of channels. The total number of patches is denoted as $N_P = H \times W/P^2$. For self-attention, each patch is projected into $Q, K, V$ with a set of weights $W_Q, W_K, W_V$ respectively. The self-attention \cite{vaswani2017attention} is computed as follows,
 \begin{equation}
     \text{Attention}(Q,K,V) =  \text{Softmax}\left( \frac{QK^T}{\sqrt{d_k}} \right) V
 \end{equation}
where $d_k$ is the dimension of the keys/queries.

\input{suppl_tables/algo}

\section{Implementation Details}
\label{sec:impl}
In this section, we describe our analysis and benchmarking
experiments, which include the augmentation strategies,
DRI dataset creation, backbone, and optimization details

\subsection{Domain Augmentations}
\label{subsec:augs}
\noindent
To induce domain-specificity, we use five label-preserving augmentations to simulate virtual domains (Fig.\ \ref{fig:augs_fig}\ak{A}):

\noindent
\textbf{a) FDA augmentation:} We use FDA \cite{FDA} to stylize an image with a fixed style-transfer set of images \cite{style_imgs}. This is done by superimposing the amplitude spectrum of the style images onto the input image.

\noindent
\textbf{b) Weather augmentations: } We employ frost and snow augmentations \cite{img_aug_lib} to augment the input images.  

\noindent
\textbf{c) AdaIN augmentation: } In this augmentation \cite{style_imgs}, we alter the feature statistics through an instance normalization layer \cite{ulyanov2017improved} that stylizes the images using the same reference style image set as in FDA.

\noindent
\textbf{d) Cartoon augmentation: } We employ cartoonization-based augmentations \cite{img_aug_lib} to convert inputs to cartoon-like images with reduced texture.

\noindent
\textbf{e) Style augmentation: } 
We use stylization from Jackson et al.\ \cite{jackson2019style}. No controllable parameters are available and style is chosen without a reference style image.

\subsection{DRI Dataset Extraction}
\label{subsec:dri_ext}

The Domain-Representative Inputs (DRI) are created using augmentations as shown in Fig.\ \ref{fig:augs_fig}. An input image is first augmented to simulate a virtual domain. Note that only one augmentation is used at a time. After this, the image is shuffled across patches to obtain a DRI image. The extent of patch shuffling is done such that the domain information is still intact, however the task-label information is lost. Following prior works \cite{mitsuzumi2021generalized}, we use normalized mutual information (NMI) to assess the consistency between the feature clusters formed by a self-supervised learning algorithm on the transformed images and the class/domain labels (see Fig.\ \ref{fig:augs_fig}\ak{C}). To obtain NMI, the training images are first subjected to the class-destructive transformation to produce DRI images, and these images are then subjected to self-supervised learning to produce class and domain invariant features. In order to assign a domain or class label to each cluster, we finally apply clustering to the learned features. For the self-supervised learning, we employ SimCLR \cite{simclr} on the DRI images and apply Gaussian mixture-based clustering to the learned features to obtain either domain or class labels for domain-NMI and class-NMI respectively.

From Figure \ref{fig:augs_fig}, we see that the Domain NMI rises within a certain range as the number of grid partitions increases, whereas the Class-NMI sharply declines. These results demonstrate that domain-specific features can be learned by using an appropriate grid partition size. Hence, for all our experiments, we have used a grid shuffling size of $4\times 4$ for representing DRI inputs.

\input{suppl_tables/ssda_dn}

\subsection{Domain-specificity disentanglement criterion}
\label{sub:dom_spec_crit}
As discussed in section 3.3 (main) paper we define the a domain-specificity disentanglement criterion based on three parameters: $\gamma_{cls}$: intra-class, inter-domain similarity, $\gamma_{dom}$: intra-domain inter-class and $\gamma_{all}$ denotes the inter-class, inter-domain similarity. We define the criterion of domain-specificity disentanglement as follows: 
\begin{equation}
    \gamma_{dom} = \expectation_{\mathcal{D}_s \cup \mathcal{D}_t}\mathcal{D}_c(z_{c_1}, z_{c_2}) \text{, where } y_{c_{1}} \neq y_{c_{2}},
    y_{d_{1}} = y_{d_{2}}
\end{equation}
\begin{equation}
    \gamma_{cls} = \expectation_{\mathcal{D}_s \cup \mathcal{D}_t}\mathcal{D}_c(z_{c_1}, z_{c_2}) \text{, where } y_{c_{1}} = y_{c_{2}},  
    y_{d_{1}} \neq y_{d_{2}}
\end{equation}
\begin{equation}
    \gamma_{dom} = \expectation_{\mathcal{D}_s \cup \mathcal{D}_t}\mathcal{D}_c(z_{c_1}, z_{c_2}) \text{, where } y_{c_{1}} \neq y_{c_{2}},  
    y_{d_{1}} \neq y_{d_{2}}
\end{equation}
\noindent
where $\mathcal{D}_c(z_{c_1}, z_{c_2})$ denotes cosine similarity between the class-token features of two inputs $x_1, x_2$ with corresponding class labels $y_{c_1}, y_{c_2}$ and domain labels $y_{d_1}, y_{d_2}$.

\vspace{1mm}
\noindent
\textbf{How to choose the threshold $\tau$?} We empirically evaluated the metrics $\gamma_{cls}$, $\gamma_{dom}$ and $\gamma_{all}$ in Table 6 (main paper) and found that task-specificity $\gamma_{cls}$ and domain-specificity $\gamma_{dom}$ are closer for DSiT (\textit{Ours}) than the SHOT-B baseline. Based on our observations, we choose a threshold of $0.05$.

\input{suppl_tables/vendor_side}

\subsection{Experimental settings}
\label{sub:exp_set}
\textbf{Backbone details.} For our experiments, we use DeiT-Base \cite{touvron2021training} which has 86M parameters, pretrained on ImageNet-1k dataset. DeiT-Base architecture consists of 12 layers, where each layer consists of multi-head self-attention with 12 heads. The input to the transformer is an RGB image that is divided into $16\times 16$-sized patches. Therefore, $P=16$ and $N_P=14$ for all our experiments. DeiT-B contains an additional distillation token, however, the rest of the architecture is the same as a ViT-B backbone. \par

\textbf{Optimization details.} For optimizing the training objectives, we use Stochastic Gradient Descent (SGD) with a momentum of $0.9$, and a weight decay ratio of $1 \times 10^{-4}$. The learning rate is set to $5 \times 10^{-3}$ for fine-tuning the domain classifier on the target domain. For the Goal Task training, we use a learning rate of $8 \times 10^{-3}$ for OfficeHome and VisDA, $8 \times 10^{-2}$ for Office-31, and $2 \times 10^{-3}$ for DomainNet. The Goal Task Training and Domain-specific disentanglement Training for the vendor-side source domain are done for $20$ epochs, of which $10$ are used for warm-up with a warm-up factor of $0.01$. In the client-side target adaptation, the goal task training (\emph{task classifier training}) is carried out for $2$ epochs, followed by the domain specificity disentanglement (\emph{domain classifier training}) until a domain classification accuracy of $80\%$ is achieved. These two steps are carried out alternatively for an effective 40 epochs of task-classifier training, the same as CDTrans \cite{xu2021cdtrans}. We use an NVIDIA RTX A5000 GPU with 64GB RAM and 24GB GPU memory to train our models. Our code takes a total training time of approximately 5 hours for Target adaptation training for the Office-Home dataset.

\section{Additional Experimental Results}
\label{sub:addn_results}

\subsection{Extended Comparisons}
\label{sub:ext_comp}

\noindent
\textbf{Comparisons on single-source domain adaptation (SSDA): } In Tables \ref{tab:ssda_dnet}, we show additional comparisons for our method with existing SSDA works on the DomainNet benchmark. We achieve significant improvements over existing works, especially on CDTrans \cite{xu2021cdtrans} despite it being a non-source-free method. It is worth noting that CDTrans uses the entire domain during the training and evaluation steps, while we train on the \emph{train} split and evaluate on the \emph{test} split, same as SSRT \cite{sun2022safe}.

\input{suppl_tables/so_comp}

\textbf{Multi-target domain adaptation (MTDA):} In Table \ref{tab:mtda_oh}, we provide a quantitative comparison with the prior arts for multi-target domain adaptation on Office-Home. The performance improvement is quite prominent (+2.0\%) over the source-free prior art (SHOT-B), and the proposed approach also yields comparable performance to non-source-free prior arts, D-CGCT and CDAN+DCL \cite{D-CGCT}, which mainly focus on domain invariant features.

\subsection{Vendor-side DSiT Performance }
\label{sub:vend_perf}
Our DSiT approach incorporates a novel Domain-Specificity Training (DST) that improves the vendor-side performance over the standard source-only baseline (shown in Table \ref{tab:vendor-side}). Further, we observe significant gains from vendor to client-side (4.5\% and 4.3\% for Office Home and DomainNet, Table \ref{tab:vendor-side}). This shows that our vendor-side DST positively aids client-side DST. 

\input{suppl_tables/vit_s}
\input{tables/mtda_oh}

\subsection{Performance in a Model Adaptation Setting}
\label{sub:adap_set}
Our DSiT approach works well even for a model adaptation setting, where we perform DST only on client-side without any specialized training on the vendor-side (\#2, Table \ref{tab:so_comp}). However, we get the best results when DST is done on both vendor and client-side (\#4, Table \ref{tab:so_comp}). We also observe that SHOT target adaptation (TA) with our vendor-side DSiT model (\#2) gives the same performance as the baseline (\#1). This indicates that our proposed TA (\#4) is able to better leverage our vendor-side model to yield improved adaptation performance.

\input{suppl_tables/train_time}

\subsection{Performance on different backbones}
\label{sub:backbones}
We report results in Table \ref{tab:vit-s} for DeiT-S backbone (with 22M parameters) pre-trained on ImageNet and observe that {our approach improves over the baseline SHOT-S baseline by {1.2\%}}. Note that ``-S" denotes Small. We also report the results over ViT-B backbone which is trained on ImageNet-21K dataset. Over ViT-B, our approach shows an improvement of {1.0\%} over the SHOT baseline.

\input{suppl_tables/a_dist}
\input{suppl_tables/task_epochs}
\input{suppl_tables/significance}

\subsection{Experimental analysis for augmentations}
\label{sub:augs_analysis}
In Table \ref{tab:a-dist augs}, we show the $\mathcal{A}$-distance (domain-gap) between augmented source and target domains on Office-Home using the class token of a source-trained DeiT-B. The domain gap for FDA is lower than the original source-target while it is higher for AdaIN. This validates Fig. 4 (main paper) which illustrated that augmented domains may be closer or farther than original domains.

\input{suppl_tables/loss_abl}

\subsection{Sensitivity Analysis of Alternate Training}
\label{sub:sens}
In our approach, we perform alternate rounds of training of the domain and the task classifier. We usually train the task classifier for a few epochs, followed by the domain classifier training. In this analysis, we vary the number of epochs of task classifier training from 1 to 5 epochs in an alternate round and observe its impact on task accuracy. Table \ref{tab:sensitivity_analysis} shows that the task accuracy increases at 2 epochs and is maximum at 3 epochs of training. In all our experiments, we report results with 2 epochs of task classifier training.

\subsection{Training time comparisons} 
\label{sub:train_time}
We provide detailed training and inference time comparisons of our method with SHOT-B in Table \ref{tab:training_time}. The DST training time is higher due to augmented images being computed at training time, which can be easily reduced by loading pre-computed augmented images. We point out that the inference time remains the same, highlighting the fact that the same DeiT-Base architecture is used for both methods.

\subsection{Statistical Significance}
\label{sub:stat_sig}
We report mean and standard deviation over 3 runs for three Office-Home settings in Table \ref{tab:mean_std}. We observe that the standard deviation ($0.1$ to $0.3$) is very low \wrt to our gains ($\sim$2\%) over SHOT-B.

\subsection{Effect of target adaptation losses}
\label{sub:abl_ta}
We perform an ablation study on 4 settings of the Office-Home dataset to analyze the influence of each component of the target adaptation objective described in Section \ref{sec:shot_loss}, and present the results in Table \ref{tab:ablation_loss}. Target adaptation with the entropy loss $\mathcal{L}_{im}$ alone shows sub-optimal results, even when compared to the source-trained baselines, which is also observed by \cite{SHOT}. Adding the diversity loss $\mathcal{L}_{div}$ shows comparatively better performance, indicating that balancing the classifier's predictions across all classes is essential.  Lastly, the self-supervised pseudo-labeling \textit{PL} also improves the performance further, demonstrating its importance towards the client-side adaptation.

\begin{figure}
\centering
       \includegraphics[width=0.8\linewidth]{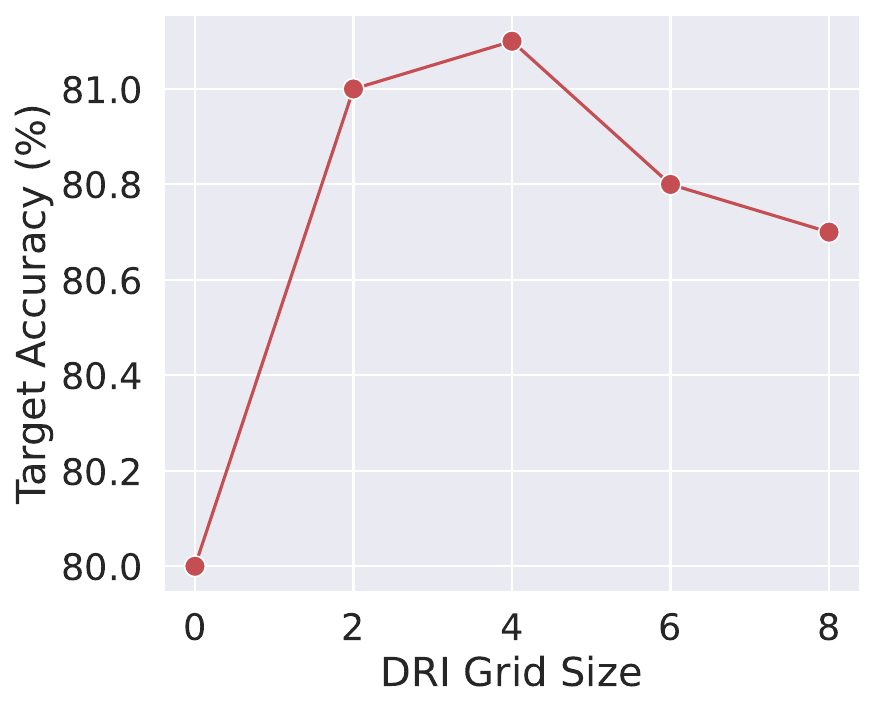}
       \caption{Sensitivity analysis on DRI grid-size for Single-Source DA on four settings of Office-Home.}
      \label{fig:dri_size}
      \vspace{-4mm}
\end{figure}

\subsection{Effect of DRI grid-size}
\label{sub:dri_size}
Here, we study the effect of varying the DRI grid size for the domain classifier training to determine its influence on the target accuracy (Figure \ref{fig:dri_size}). The target accuracy gradually increases upon increasing the grid size from 1 to 4, with the best results being observed at grid size 4. Beyond this point, the performance begins to drop, which can be attributed to the excessive destruction of information caused by over-partitioning of the images. To achieve a balance between the effect of the task-destructive transformation and its impact on the target accuracy, we use a grid size of $4 \times 4$ for all our experiments.

{\small
\bibliographystyle{ieee_fullname}
\bibliography{main}
}

\end{document}

%% file: tables/ssda_oh.tex
\begin{table*}[t]
    \centering
    \setlength{\tabcolsep}{2.0pt}
    \caption{Single-Source Domain Adaptation (SSDA) on Office-Home benchmark. SF indicates \textit{source-free} adaptation. ResNet-based methods (top) and Transformer-based methods (bottom). 
    * indicates results taken from CDTrans \cite{xu2021cdtrans}. \textbf{Bold} numbers indicate the best results among SFDA methods. \textbf{\textcolor{ForestGreen}{Green}} results indicate our method's improvements over the corresponding methods.
    }
    \label{tab:ssda_oh}
    \resizebox{\linewidth}{!}{
        \begin{tabular}{lccccccccccccccl}
            \toprule
            \multirow{2}{30pt}{\centering Method} & \multirow{2}{*}{\centering SF} & &\multicolumn{13}{c}{\textbf{Office-Home}} \\
            \cmidrule{4-16}
            && & { {Ar}$\shortrightarrow${Cl}} & { {Ar}$\shortrightarrow${Pr}} & {{Ar}$\shortrightarrow${Rw}} & {{Cl}$\shortrightarrow${Ar}} & {{Cl}$\shortrightarrow${Pr}} & {{Cl}$\shortrightarrow${Rw}} & {{Pr}$\shortrightarrow${Ar}} & {{Pr}$\shortrightarrow${Cl}} & {{Pr}$\shortrightarrow${Rw}} & {{Rw}$\shortrightarrow${Ar}} & {{Rw}$\shortrightarrow${Cl}} & {{Rw}$\shortrightarrow${Pr}} & Avg \\
            \midrule
            ResNet-50 \cite{he2016deep_resnet} & \xmark && 34.9 & 50.0 & 58.0 & 37.4 & 41.9 & 46.2 & 38.5 & 31.2 & 60.4 & 53.9 & 41.2 & 59.9 & 46.1 \\
            SENTRY\cite{SENTRY} & \xmark && 61.8 & 77.4 & 80.1 & 66.3 & 71.6 & 74.7 & 66.8 & 63.0 & 80.9 & 74.0 & 66.3 & 84.1 & 72.2 \\
            SCDA~\cite{SCDA} & \xmark && 60.7 & 76.4 & 82.8 & 69.8 & 77.5 & 78.4 & 68.9 & 59.0 & 82.7 & 74.9 & 61.8 & 84.5 & 73.1 \\
            
            ${\text{A}^{2}\text{Net}}$ \cite{A2Net} &\cmark& &  58.4 & 79.0 & {82.4} & 67.5 & 79.3 & 78.9 & {68.0} & 56.2 & 82.9 & {74.1} & 60.5 & 85.0 & 72.8 \\
            GSFDA~\cite{GSFDA} & \cmark && 57.9 & 78.6 & 81.0 & 66.7 & 77.2 & 77.2 & 65.6 & 56.0 & 82.2 & 72.0 & 57.8 & 83.4 & 71.3 \\
            CPGA~\cite{CPGA} & \cmark && {59.3} & 78.1 & 79.8 & 65.4 & 75.5 & 76.4 & 65.7 & {58.0} & 81.0 & 72.0 & {64.4} & 83.3 & 71.6 \\
            {NRC}~\cite{NRC} &\cmark& &{57.7} 	& {80.3} 	&{82.0} 	&{68.1} 	&{79.8} 	&{78.6} 	&{65.3} 	&{56.4} 	&{83.0} 	&71.0	&{58.6} &{85.6} 	&{72.2} \\
            SHOT~ \cite{SHOT} &\cmark& &{57.1} & {78.1} & 81.5 & {68.0} & {78.2} & {78.1} & {67.4} & 54.9 & {82.2} & 73.3 & 58.8 & {84.3} & {71.8}  \\
            SHOT++ \cite{SHOT++} & \cmark && 57.9 & 79.7 & 82.5 & {68.5} & 79.6 & 79.3 & {68.5} & 57.0 & {83.0} & 73.7 & 60.7 & 84.9 & 73.0 \\
            \midrule
            TVT \cite{yang2021tvt} & \xmark && 74.8 & 86.8 & 89.4 & 82.7 & 87.9 & 88.2 & 79.8 & 71.9 & 90.1 & 85.4 & 74.6 & 90.5 & 83.5 \\
            SSRT-B \cite{sun2022safe}& \xmark && 75.1 & 88.9 & 91.0 & 85.1 & 88.2 & 89.9 & 85.0 & 74.2 & 91.2 & 85.7 & 78.5 & 91.7 & 85.4 \\
            CDTrans \cite{xu2021cdtrans} & \xmark && 68.8 & 85.0 & 86.9 & 81.5 & 87.1 & 87.3 & 79.6 & 63.3 & 88.2 & 82.0 & 66.0 & 90.6 & 80.5 \\
            \rowcolor{Gray} SHOT-B* & \cmark && 67.1  & 83.5  & 85.5  & 76.6  & 83.4  & 83.7  & 76.3  & 65.3  & 85.3  & 80.4  & 66.7 & 83.4 & {78.1} \small{\textbf{\color{ForestGreen}{(+2.4)}}}\\
            
            \rowcolor{Gray} DIPE \cite{wang2022exploring} & \cmark && 66.0 & 80.6 & 85.6 & 77.1 & 83.5 & 83.4 & 75.3 & 63.3 & 85.1 & 81.6 & 67.7 & 89.6 & 78.2 \small{\textbf{\color{ForestGreen}{(+2.3)}}} \\
            
             \rowcolor{Gray} Mixup \cite{kundu2022balancing} & \cmark && 65.3 & 82.1 & 86.5 & 77.3 & 81.7 & 82.4 & 77.1 & 65.7 & 84.6 & 81.2 & \textbf{70.1} & 88.3 & 78.5 \small{\textbf{\color{ForestGreen}{(+2.0)}}} \\

            \rowcolor{Gray} DSiT-B (\textit{Ours}) & \cmark && \textbf{69.2} & \textbf{83.5} & \textbf{87.3} & \textbf{80.7} & \textbf{86.1} & \textbf{86.2} & \textbf{77.9} & \textbf{67.9} & \textbf{86.6} & \textbf{82.4} & 68.3 & \textbf{89.8} & \textbf{80.5} \\
            \bottomrule
            \end{tabular} 
        }
\end{table*}

%% file: tables/ssda_o31.tex
\begin{table*}[t]
    \centering
    \setlength{\tabcolsep}{8.7pt}
    \renewcommand{\arraystretch}{0.95}
    \caption{Single-Source Domain Adaptation (SSDA) on Office-31 and VisDA benchmarks. SF indicates \textit{source-free} adaptation.
    ResNet-based methods (top) and Transformer-based methods (bottom). 
    * indicates results taken from CDTrans \cite{xu2021cdtrans}. \textbf{Bold} numbers indicate the best results among SFDA methods. \textbf{\textcolor{ForestGreen}{Green}} results indicate our method's improvements over the corresponding methods.
    }
    \label{tab:ssda_o31}
    \resizebox{0.95\linewidth}{!}{
        \setlength{\extrarowheight}{1pt}
        \begin{tabular}{lccccrccclll}
            \toprule
            \multirow{2}{30pt}{\centering Method} & \multirow{2}{*}{\centering SF} && \multicolumn{7}{c}{\textbf{Office-31}} && \textbf{VisDA}\\
            \cmidrule{4-10} \cmidrule{12-12}
            &&& A$\shortrightarrow$W & D$\shortrightarrow$W & W$\shortrightarrow$D & A$\shortrightarrow$D & D$\shortrightarrow$A & W$\shortrightarrow$A & Avg. && S$\shortrightarrow$R \\
            \midrule
            ResNet-50 \cite{he2016deep_resnet} & \xmark & & 68.9 & 68.4 & 62.5 & 96.7 & 60.7 & 99.3 & 76.1 & & 52.4\\
            CAN~\cite{CAN} & \xmark& &  94.5 & 99.1 & 99.8 & 95.0 & 78.0 & 77.0 & 90.6 & &87.2 \\
            FixBi~\cite{FixBi} & \xmark& & 96.1 & 99.3 & 100.0 & 95.0 & 78.7 & 79.4 & 91.4 && 87.2 \\
            CDAN+RADA~\cite{RADA} & \xmark && 96.2 & 99.3 & 100.0 & 96.1 & 77.5 & 77.4 & 91.1 && 76.3\\
            CPGA~\cite{CPGA} & \cmark && 94.1 & 98.4 & 99.8 & 94.4 & 76.0 & 76.6 & 89.9 && 84.1\\
            HCL~\cite{HCL} & \cmark && 91.3 & 98.2 & 100.0 & 90.8 & 72.7 & 72.7 & 87.6 && 83.5\\
            VDM-DA~\cite{VDM-DA}  & \cmark && 94.1 & 98.0 & 100.0 & 93.2 & 75.8 & 77.1 & 89.7 && 85.1\\
            ${\text{A}^{2}\text{Net}}$~ \cite{A2Net} & \cmark && {94.0} & 99.2 & 100.0 & 94.5 & 76.7 & 76.1 & 90.1 && 84.3\\
            NRC~\cite{NRC} & \cmark && 90.8 & 99.0 & 100.0 & {96.0} & 75.3 & 75.0 & 89.4 && 85.9\\
            SHOT~ \cite{SHOT}& \cmark& & 90.1 &  98.4 & {99.9} & {94.0} & 74.7 & 74.3 & {88.6} && 82.9\\
            SHOT++~\cite{SHOT++} & \cmark && 90.4 & 98.7 & 99.9 & 94.3 & 76.2 & 75.8 & 89.2 && {87.3}\\
            \midrule
            TVT \cite{yang2021tvt}& \xmark & & 96.4 & 99.4 & 100.0 & 96.4 & 84.9 & 86.1 & 93.8 && 83.9\\
            CGDM-B* \cite{du2021cross} & \xmark & & 95.3 & 97.6 & 99.8 & 94.6 & 78.8 & 81.2 & 91.2 && 82.3\\
            CD-Trans \cite{xu2021cdtrans} & \xmark & & 96.7 & 99.0 & 100.0 & 97.0 & 81.1 & 81.9 & 92.6 && 88.4\\
            SSRT-B \cite{sun2022safe} & \xmark & & 97.7 & 99.2 & 100.0 & 98.6 & 83.5 & 82.2 & 93.5 && 88.7\\
            \rowcolor{Gray} DIPE \cite{wang2022exploring} & \cmark & & 95.5 & 98.5 & 100.0 & 94.8 & 77.5 & 77.1 & 90.5 \small{\textbf{\color{ForestGreen}{(+2.5)}}} && 82.8 \small{\textbf{\color{ForestGreen}{(+4.8)}}} \\

            \rowcolor{Gray} Mixup \cite{kundu2022balancing} & \cmark & & 96.1 & 98.6 & 100.0 & 95.4 & 80.2 & 80.1 & 91.7 \small{\textbf{\color{ForestGreen}{(+1.3)}}} && 86.3 \small{\textbf{\color{ForestGreen}{(+1.3)}}} \\
            
            \rowcolor{Gray} SHOT-B* & \cmark & & 94.3 & 99.0 & 100.0 & 95.3 & 79.4 & 80.2 & 91.4 \small{\textbf{\color{ForestGreen}{(+1.6)}}} && 85.9 \small{\textbf{\color{ForestGreen}{(+1.7)}}} \\
            \rowcolor{Gray} DSiT-B (\textit{Ours}) & \cmark & & \textbf{97.2} & \textbf{99.1} & \textbf{100.0} & \textbf{98.0} & \textbf{81.7} & \textbf{81.8} & \textbf{93.0} && \textbf{87.6} \\
            \bottomrule
            \end{tabular} 
        }
\end{table*}

%% file: tables/ssda_dn.tex
\begin{table*}[t]
\centering
\setlength{\tabcolsep}{2.2pt}
\caption{Single-Source Domain Adaptation (SSDA) results on the DomainNet dataset. * indicates results taken from \cite{sun2022safe}. NSF indicates Non-Source-Free, SF indicates Source-Free and SO indicates Source-Only.}
\label{tab:ssda_dn}
\resizebox{\textwidth}{!}{
\begin{tabular}[t]{ccc}
        
        \begin{tabular}[t]{|C{2cm}|ccccccc|}
            \hline
            \begin{tabular}{c}
            \textbf{SCDA} \\ \cite{MDD} \textbf{(NSF)} \end{tabular} & clp & inf & pnt & qdr & rel & skt & Avg. \\
            \hline
            clp  & -    & 20.4 & 43.3 & 15.2 & 59.3 & 46.5 & 36.9 \\
            inf  & 32.7 & -    & 34.5 & 6.3  & 47.6 & 29.2 & 30.1 \\
            pnt  & 46.4 & 19.9 & -    & 8.1  & 58.8 & 42.9 & 35.2 \\
            qdr  & 31.1 & 6.6  & 18.0 & -    & 28.8 & 22.0 & 21.3 \\
            rel  & 55.5 & 23.7 & 52.9 & 9.5  & -    & 45.2 & 37.4 \\
            skt  & 55.8 & 20.1 & 46.5 & 15.0 & 56.7 & -    & 38.8 \\
            Avg. & 44.3 & 18.1 & 39.0 & 10.8 & 50.2 & 37.2 & \cellcolor{Gray} \textbf{33.3} \\
            \hline
        \end{tabular} &
        \begin{tabular}[t]{|C{2cm}|ccccccc|}
            \hline
            \begin{tabular}{c} \textbf{DeiT-B} \\  \cite{touvron2021training} \textbf{(SO)} \end{tabular} & clp & inf & pnt & qdr & rel & skt & Avg. \\
            \hline
            clp  & -    & 24.3 & 49.6 & 15.8 & 65.3 & 52.1 & 41.4 \\
            inf  & 45.9 & -    & 45.9 & 6.7  & 61.4 & 39.5 & 39.9 \\
            pnt  & 53.2 & 23.8 & -    & 6.5  & 66.4 & 44.7 & 38.9 \\
            qdr  & 31.9 & 6.8  & 15.4 & -    & 23.4 & 20.6 & 19.6 \\
            rel  & 59.0 & 25.8 & 56.3 & 9.16 & -    & 44.8 & 39.0 \\
            skt  & 60.6 & 20.6 & 48.4 & 16.5 & 61.2 & -    & 41.5 \\
            Avg. & 50.1 & 20.3 & 43.1 & 10.9 & 55.5 & 40.3 & \cellcolor{Gray} \textbf{36.7} \\
            \hline
        \end{tabular} &
        \begin{tabular}[t]{|C{2cm}|ccccccc|}
            \hline
            \begin{tabular}{c} \textbf{SHOT-B} \\  \cite{SHOT} \textbf{(SF)}\end{tabular} & clp & inf & pnt & qdr & rel & skt & Avg. \\
            \hline
            clp  & -    & 27.0 & 49.7 & 16.5 & 65.4 & 53.2 & 46.1 \\
            inf  & 46.4 & -    & 45.9 & 7.4  & 60.6 & 40.1 & 40.1 \\
            pnt  & 54.6 & 25.7 & -    & 8.1  & 66.3 & 49.0 & 40.7 \\
            qdr  & 33.3 & 6.8  & 15.5 & -    & 23.8 & 24.0 & 20.7 \\
            rel  & 59.3 & 28.1 & 57.4 & 9.0  & -    & 47.3 & 40.2 \\
            skt  & 64.0 & 26.5 & 55.0 & 18.2 & 63.8 & -    & 45.5 \\
            Avg. & 51.5 & 26.6 & 44.7 & 11.8 & 56.0 & 42.7 & \cellcolor{Gray} \textbf{38.9} \\
            \hline
        \end{tabular} \\
        \begin{tabular}[t]{|C{2cm}|ccccccc|}
            \hline
            \begin{tabular}{c} \textbf{CDTrans$^*$} \\  \cite{xu2021cdtrans} \textbf{(NSF)} \end{tabular} & clp & inf & pnt & qdr & rel & skt & Avg. \\
            \hline
            clp  & -    & 27.9 & 57.6 & 27.9 & 73.0 & 58.8 & 49.0 \\
            inf  & 58.6 & -    & 53.4 & 9.6  & 71.1 & 47.6 & 48.1 \\
            pnt  & 60.7 & 24.0 & -    & 13.0 & 69.8 & 49.6 & 43.4 \\
            qdr  & 2.9  & 0.4  & 0.3  & -    & 0.7  & 4.7  & 1.8  \\
            rel  & 49.3 & 18.7 & 47.8 & 9.4  & -    & 33.5 & 31.7 \\
            skt  & 66.8 & 23.7 & 54.6 & 27.5 & 68.0 & -    & 48.1 \\
            Avg. & 47.7 & 18.9 & 42.7 & 17.5 & 56.5 & 38.8 & \cellcolor{Gray} \textbf{37.0} \\
            \hline
        \end{tabular} &
        \begin{tabular}[t]{|C{2cm}|ccccccc|}
            \hline
            \begin{tabular}{c} \textbf{SSRT-B$^*$} \\  \cite{sun2022safe} \textbf{(NSF)}\end{tabular} & clp & inf & pnt & qdr & rel & skt & Avg. \\
            \hline
            clp  & -    & 33.8 & 60.2 & 19.4 & 75.8 & 59.8 & 49.8 \\
            inf  & 55.5 & -    & 54.0 & 9.0  & 68.2 & 44.7 & 46.3 \\
            pnt  & 61.7 & 28.5 & -    & 8.4  & 71.4 & 55.2 & 45.0 \\
            qdr  & 42.5 & 8.8  & 24.2 & -    & 37.6 & 33.6 & 29.3 \\
            rel  & 69.9 & 37.1 & 66.0 & 10.1 & -    & 58.9 & 48.4 \\
            skt  & 70.6 & 32.8 & 62.2 & 21.7 & 73.2 & -    & 52.1 \\
            Avg. & 60.0 & 28.2 & 53.3 & 13.7 & 65.3 & 50.4 & \cellcolor{Gray} \textbf{45.2} \\
            \hline
        \end{tabular} &
        \begin{tabular}[t]{|C{2cm}|ccccccc|}
            \hline
            \begin{tabular}{c} \textbf{DSiT} \\  \textbf{(\textit{Ours}) (SF)} \end{tabular} & clp & inf & pnt & qdr & rel & skt & Avg. \\
            \hline
            clp  & -    & 27.2 & 51.8 & 23.1 & 70.2 & 54.7 & 45.4 \\
            inf  & 52.3 & -    & 48.8 & 12.8 & 68.3 & 44.2 & 45.3 \\
            pnt  & 59.2 & 26.1 & -    & 14.5 & 71.5 & 51.4 & 44.5 \\
            qdr  & 38.1 & 8.3  & 21.2 & -    & 37.2 & 27.6 & 26.5 \\
            rel  & 60.4 & 28.0 & 57.8 & 13.1 & -    & 49.7 & 41.8 \\
            skt  & 66.3 & 27.5 & 56.0 & 24.4 & 70.2 & -    & 48.9 \\
            Avg. & 55.3 & 23.4 & 47.1 & 17.6 & 63.5 & 45.5 & \cellcolor{Gray} \textbf{42.1} \\
            \hline
        \end{tabular}
    \end{tabular}
    }
\end{table*}

%% file: tables/msda_oh.tex
\begin{table}[t]
    \centering
    \setlength{\tabcolsep}{6pt}
    \setlength{\extrarowheight}{1pt}
    \caption{Multi-Source DA (MSDA) on Office-Home benchmark. SF indicates \textit{source-free} adaptation.
    ResNet-based methods (top) and Transformer-based methods (bottom).
    }
    \label{tab:msda_oh_dn}
    \resizebox{0.9\linewidth}{!}{
        \setlength{\extrarowheight}{1pt}
        \begin{tabular}{lcccccl}
            \toprule
            \multirow{2}{30pt}{\centering Method} & \multirow{2}{*}{\centering SF} & \multicolumn{5}{c}{\textbf{Office-Home}} \\
            \cmidrule{3-7}
            && $\shortrightarrow$Ar & $\shortrightarrow$Cl & $\shortrightarrow$Pr & $\shortrightarrow$Rw & Avg. \\
            \midrule
            Source-combine & \xmark & 58.0   & 57.3 & 74.2 & 77.9 & 66.9 \\
            WDA \cite{WAMDA} & \xmark & 71.9 & 61.4 & 84.1 & 82.3 & 74.9 \\
            SImpAl \cite{SImpAl} & \xmark & 70.8 & 56.3 & 80.2 & 81.5 & 72.2 \\
            CMSDA \cite{CMSDA} & \xmark & 71.5 & 67.7 & 84.1 & 82.9 & 76.6 \\
            DECIS \cite{DECISION} & \cmark & 74.5 & 59.4 & 84.4 & 83.6 & 75.5 \\
            SHOT++ \cite{SHOT++} & \cmark & 73.1 & 61.3 & 84.3 & 84.0 & 75.7 \\
            CAiDA \cite{CAiDA} & \cmark & 75.2 & 60.5 & 84.7 & 84.2 & 76.2 \\
            SHOT \cite{SHOT} & \cmark & 72.2 & 59.3 & 82.8 & 82.9 & 74.3 \\
            \midrule
            \rowcolor{Gray} SHOT-B* & \cmark & 83.9 & 71.8 & 89.7 & 89.4 & 83.7 \\
            \rowcolor{Gray} DSiT-B (\textit{Ours}) & \cmark & \textbf{84.4} & \textbf{73.8} & \textbf{90.7} & \textbf{89.7} & \textbf{84.7} \textbf{\color{ForestGreen}{(+1.0)}}\\
            
            \bottomrule
            \end{tabular} 
        }
\end{table}

%% file: tables/dri_ablation.tex
\begin{table}[t]
\centering
\setlength{\tabcolsep}{6pt}
\caption{Ablation study for DRI as a general augmentation on Office-Home SSDA benchmark with the DeiT-B backbone.}
\vspace{0.5mm}
\label{tab:dri_abl}
\resizebox{\linewidth}{!}{
\begin{tabular}{lccccc}
\toprule
{Method} & Ar$\shortrightarrow$Cl & Cl$\shortrightarrow$Pr & Pr$\shortrightarrow$Rw & Rw$\shortrightarrow$Ar & Avg. \\
\midrule

SHOT-B \cite{SHOT}                & 67.1 & 83.4 & 85.3 & 80.4 & 79.1 \\
SHOT-B + DRI          & 65.8 & 79.9 & 84.2 & 79.6 & 77.4 \\
\midrule
CDTrans \cite{xu2021cdtrans}            & 68.8 & 87.1 & 88.2 & 82.0 & 81.5 \\
CDTrans + DRI       & 59.7 & 81.6 & 84.2 & 81.5 & 76.8 \\
\bottomrule
\end{tabular}
}
\end{table}

%% file: tables/dsit_ablation.tex
\begin{table}[t]
    \centering
    \setlength{\tabcolsep}{9pt}
    \setlength{\extrarowheight}{2pt}
    \caption{Ablation study for various stages of training on Office-Home SSDA benchmark with the DeiT-B backbone (\textit{average over 4 settings}). SO: Source-Only, DST: Domain-Specificity Training} 
    \label{tab:ablation}
    \resizebox{0.95\columnwidth}{!}{
        \begin{tabular}{clcccl}
        \toprule
        Training Phase & Model & \# & DST & DRI & Avg. \\
        \midrule
        \multirow{3}{*}{Vendor-side} & SO & 1 &\xmark & \xmark &  74.8 \\ [1pt]
                                     & \cellcolor{Gray} & \cellcolor{Gray}2 & \cellcolor{Gray}\cmark & \cellcolor{Gray}\xmark & \cellcolor{Gray}75.6 \small{\textbf{\color{ForestGreen}{(+0.8)}}} \\ [1pt]
                                     & \cellcolor{Gray}\multirow{-2}{*}{DSiT} & \cellcolor{Gray}3 & \cellcolor{Gray}\cmark & \cellcolor{Gray}\cmark & \cellcolor{Gray}76.6 \small{\textbf{\color{ForestGreen}{(+1.8)}}} \\ [1pt]
        \hline
         \multirow{3}{*}{Client-side} & SHOT & 4 & \xmark & \xmark & 79.0 \\ [1pt]
                                      & \cellcolor{Gray} & \cellcolor{Gray}5 & \cellcolor{Gray}\cmark & \cellcolor{Gray}\xmark & \cellcolor{Gray}80.0 \small{\textbf{\color{ForestGreen}{(+1.0)}}} \\ [1pt]
                                      & \cellcolor{Gray}\multirow{-2}{*}{DSiT} & \cellcolor{Gray}6 & \cellcolor{Gray}\cmark & \cellcolor{Gray}\cmark & \cellcolor{Gray}\textbf{81.1} \small{\textbf{\color{ForestGreen}{(+2.1)}}} \\    
        \bottomrule
        \end{tabular}
        }
\end{table}

%% file: tables/dom_spec_crit.tex
\begin{table}[t]
    \centering
    \setlength{\tabcolsep}{12pt}
    \setlength{\extrarowheight}{1pt}
    \caption{Empirical evaluation of similarity metrics shows that domain-specificity $\gamma_{dom}$ and task-specificity $\gamma_{cls}$ are closer for our DSiT along with better DA performance, indicating better disentanglement than the baseline.} 
    \label{tab:defn1_analysis}
    \resizebox{\columnwidth}{!}{
        \begin{tabular}{lcccc}
            \toprule
            Office-Home & $\gamma_{cls}$ & $\gamma_{dom}$ & $\gamma_{all}$ & SSDA \\
            \midrule
            SHOT-B \cite{SHOT} & 0.84 & 0.74 & 0.73 & 78.1 \\
            DSiT-B (\textit{Ours}) & 0.81 & 0.78 & 0.71 & \textbf{80.5} \\
            \bottomrule
        \end{tabular} 
        }
\vspace{-2mm}
\end{table} 

%% file: suppl_tables/notations.tex
\begin{table}[ht]
    \centering
    \caption{List of all the notations used throughout the paper.}
    \label{sup:tab:notations}
    \vspace{1mm}
    \setlength{\tabcolsep}{11pt}
    \resizebox{0.9\columnwidth}{!}{
        \begin{tabular}{lcl}
        \toprule
        
         & \multirow{1}{*}{Symbol} & \multirow{1}{*}{Description} \\
         \midrule

         \multirow{3}{*}{\rotatebox[origin=c]{90}{Models}} & $h$ & Backbone feature extractor  \\
         & $f_g$ & Goal task classifier \\
         & $f_d$ & Domain classifier \\
         \midrule

        \multirow{9}{*}{\rotatebox[origin=c]{90}{Transformers}} 
         & $z_c$ & Class token of last layer \\
         
         & $z_d$ & Domain token of last layer\\
        
         & $N_P$ & Number of patch tokens \\
         & $W_Q$ & Query weights\\
         & $W_K$ & Key weights\\
         & $W_V$ & Value weights\\
         & $\theta_Q$ & Query weights of all layers \\
         & $\theta_K$ & Key weights of all layers \\
         & $\theta_V$ & Value weights of all layers \\

        \midrule
        \multirow{9}{*}{\rotatebox[origin=c]{90}{Datasets}} & $\mathcal{D}_s$ & Labeled source dataset  \\
         & $\mathcal{D}_t$ & Unlabeled target dataset  \\
         & $\mathcal{A}_i$ & $i^\text{th}$ augmentation function \\
         & $\mathcal{D}_s^{[i]}$ & $i^\text{th}$ augmented source dataset \\
         & $\mathcal{D}_t^{[i]}$ & $i^\text{th}$ augmented target dataset \\
         & $(x_s, y_{s})$ & Labeled source sample  \\
         & $(x_{s}^{[i]}, y_s, y_d)$ & Augmented source sample  \\
         & $x_t$ & Unlabeled target sample  \\
         & $(x_{t}^{[i]}, y_d)$ & Target augmented sample \\
         \midrule 
        \multirow{5}{*}{\rotatebox[origin=c]{90}{Spaces}} 
        & $\mathcal{X}$ & Input space \\
        & $\mathcal{C}_g$ & Label set for goal task \\
        & $\mathcal{Z}_c$ & Class token feature space\\
        & $\mathcal{Z}_d$ & Domain token feature space\\
        & $\mathcal{Z}_1,\dots, \mathcal{Z}_{N_P}$ & Patch token \\

        \midrule
        \multirow{4}{*}{\rotatebox[origin=c]{90}{ Losses}} & $\mathcal{L}_{dom}$ & Domain classification loss  \\
         & $\mathcal{L}_{cls}$ & Task classification loss  \\
         & $\mathcal{L}_{im}$ & Entropy loss  \\
         & $\mathcal{L}_{div}$ & Diversity loss \\

         \midrule
        \multirow{4}{*}{\rotatebox[origin=c]{90}{ Criterion}} & $\gamma_{dom}$ & Domain specificity \\
         & $\gamma_{cls}$ & Task specificity\\
         & $\gamma_{all}$ & Inter-class-inter-domain similarity \\
         & $\tau$ & Threshold  \\
        
        \bottomrule
        \end{tabular}
        }
\end{table}

%% file: suppl_tables/algo.tex
\begin{algorithm}[!t]

\caption{DSiT Training Algorithm}
\label{algo:overall}
\begin{algorithmic}[1]
\vspace{1mm}
\Statex \underline{\textbf{Vendor-side training}}

\vspace{1mm}
\State \textbf{Input:} Let $\mathcal{D}_s$ be source data , $\mathcal{D}_{s}^{[i]}$ be augmented DRI dataset for each augmentation $\mathcal{A}_i$, ImageNet pretrained DeiT-B backbone $h$ from \cite{xu2021cdtrans}, randomly initialized goal classifier $f_g$ and randomly initialized domain classifier $f_d$.

\vspace{1mm}
\For{$iter < MaxIter$}:
\vspace{1mm}
\Statex \underline{{\textit{Goal task training}}}
\vspace{1mm}
\For{$iter < MaxTaskIters$}:
\State Sample batch from $\mathcal{D}_s$ 
\State Compute $\mathcal{L}_{cls}$ using Eq.\ 2 (main paper)
\State \textbf{update} $\theta_h\setminus\theta_Q, \theta_{f_g}$ by minimizing $\mathcal{L}_{cls}$
\EndFor

\vspace{1mm}
\Statex \underline{{\textit{Domain classifier training}}}
\vspace{1mm}
\For{$iter < MaxDomainIters$}:
\State Sample batch of DRI from $\mathcal{D}_{s}^{[i]}$
\State Compute $\mathcal{L}_{dom}$ using Eq.\ {1} (main paper)
\State \textbf{update} $\theta_{Q},\theta_{f_d}$ by minimizing $\mathcal{L}_{dom}$
\EndFor
\vspace{1mm}
\Statex \jnkc{\Comment{The two steps are carried out alternatively
}}
\EndFor

\vspace{1mm}
\Statex \underline{\textbf{Client-side training}}

\vspace{1mm}
\State \textbf{Input:} Target data $\mathcal{D}_t$, Target augmented DRI data $\mathcal{D}_t^{[i]}$, source-side pretrained backbone $h$, goal classifier $f_g$ and domain classifier $f_d$.

\vspace{1mm}
\For{$iter < MaxIter$}:
\vspace{1mm}
\Statex \underline{{\textit{Goal Task Training}}}
\vspace{1mm}
\For{$iter < MaxTaskIters$}:
\State Sample batch from $\mathcal{D}_t$
\State Compute $\mathcal{L}_{im}$ and $\mathcal{L}_{div}$ using Eq.\ \ref{sup:eqn:loss_im}, \ref{sup:eqn:loss_div} (suppl.)
\State \textbf{update} $\theta_h \setminus \theta_Q, \theta_{f_g}$ by minimizing $\mathcal{L}_{im} + \mathcal{L}_{div}$
\EndFor

\vspace{1mm}
\Statex \underline{{\textit{Domain classifier training}}}
\vspace{1mm}
\For{$iter < MaxDomainIters$}:
\State Sample batch of DRI from $\mathcal{D}_{t}^{[i]}$
\State Compute $\mathcal{L}_{dom}$ using Eq.\ {1} (main paper)
\State \textbf{update} $\theta_{Q},\theta_{f_d}$ by minimizing $\mathcal{L}_{dom}$
\EndFor
\vspace{1mm}
\Statex \jnkc{\Comment{The two steps are carried out alternatively.
}}

\EndFor

\end{algorithmic}
\end{algorithm}

%% file: suppl_tables/ssda_dn.tex
\begin{table*}[t]
\centering
\setlength{\tabcolsep}{2.2pt}
\caption{Single-Source Domain Adaptation (SSDA) results on the DomainNet dataset. * indicates results taken from \cite{sun2022safe}.}
\label{tab:ssda_dnet}
\resizebox{\textwidth}{!}{
\begin{tabular}[t]{ccc}
    
        \begin{tabular}[t]{|C{2cm}|ccccccc|}
            \hline
            \begin{tabular}{c} \textbf{ResNet-} \\ \textbf{101} \cite{he2016deep_resnet} \end{tabular} & clp & inf & pnt & qdr & rel & skt & Avg. \\
            \hline
            clp  & -    & 19.3 & 37.5 & 11.1 & 52.2 & 41.0 & 32.2 \\
            inf  & 30.2 & -    & 31.2 & 3.6  & 44.0 & 27.9 & 27.4 \\
            pnt  & 39.6 & 18.7 & -    & 4.9  & 54.5 & 36.3 & 30.8 \\
            qdr  & 7.0  & 0.9  & 1.4  & -    & 4.1  & 8.3  & 4.3  \\
            rel  & 48.4 & 22.2 & 49.4 & 6.4  & -    & 38.8 & 33.0 \\
            skt  & 46.9 & 15.4 & 37.0 & 10.9 & 47.0 & -    & 31.4 \\
            Avg. & 34.4 & 15.3 & 31.3 & 7.4  & 40.4 & 30.5 & \cellcolor{Gray} \textbf{26.6} \\
            \hline
        \end{tabular} &
        \begin{tabular}[t]{|C{2cm}|ccccccc|}
            \hline
            \begin{tabular}{c} \textbf{CDAN} \\ \cite{CDAN} \end{tabular} & clp & inf & pnt & qdr & rel & skt & Avg. \\
            \hline
            clp  & -    & 20.4 & 36.6 & 9.0  & 50.7 & 42.3 & 31.8 \\
            inf  & 27.5 & -    & 25.7 & 1.8  & 34.7 & 20.1 & 22.0 \\
            pnt  & 42.6 & 20.0 & -    & 2.5  & 55.6 & 38.5 & 31.8 \\
            qdr  & 21.0 & 4.5  & 8.1  & -    & 14.3 & 15.7 & 12.7 \\
            rel  & 51.9 & 23.3 & 50.4 & 5.4  & -    & 41.4 & 34.5 \\
            skt  & 50.8 & 20.3 & 43.0 & 2.9  & 50.8 & -    & 33.6 \\
            Avg. & 38.8 & 17.7 & 32.8 & \textcolor{white}{0}4.3  & 41.2 & 31.6 & \cellcolor{Gray} \textbf{27.7} \\
            \hline
        \end{tabular} &
        \begin{tabular}[t]{|C{2cm}|ccccccc|}
            \hline
            \begin{tabular}{c} \textbf{MIMFTL} \\  \cite{MIMFTL} \end{tabular} & clp & inf & pnt & qdr & rel & skt & Avg. \\
            \hline
            clp  & -    & 15.1 & 35.6 & 10.7 & 51.5 & 43.1 & 31.2 \\
            inf  & 32.1 & -    & 31.0 & 2.9  & 48.5 & 31.0 & 29.1 \\
            pnt  & 40.1 & 14.7 & -    & 4.2  & 55.4 & 36.8 & 30.2 \\
            qdr  & 18.8 & 3.1  & 5.0  & -    & 16.0 & 13.8 & 11.3 \\
            rel  & 48.5 & 19.0 & 47.6 & 5.8  & -    & 39.4 & 32.1 \\
            skt  & 51.7 & 16.5 & 40.3 & 12.3 & 53.5 & -    & 34.9 \\
            Avg. & 38.2 & 13.7 & 31.9 & 7.2  & 45.0 & 32.8 & \cellcolor{Gray} \textbf{28.1} \\
            \hline
        \end{tabular} \\
        \begin{tabular}[t]{|C{2cm}|ccccccc|}
            \hline
            \begin{tabular}{c} \textbf{MDD+} \\  \textbf{SCDA} \cite{MDD} \end{tabular} & clp & inf & pnt & qdr & rel & skt & Avg. \\
            \hline
            clp  & -    & 20.4 & 43.3 & 15.2 & 59.3 & 46.5 & 36.9 \\
            inf  & 32.7 & -    & 34.5 & 6.3  & 47.6 & 29.2 & 30.1 \\
            pnt  & 46.4 & 19.9 & -    & 8.1  & 58.8 & 42.9 & 35.2 \\
            qdr  & 31.1 & 6.6  & 18.0 & -    & 28.8 & 22.0 & 21.3 \\
            rel  & 55.5 & 23.7 & 52.9 & 9.5  & -    & 45.2 & 37.4 \\
            skt  & 55.8 & 20.1 & 46.5 & 15.0 & 56.7 & -    & 38.8 \\
            Avg. & 44.3 & 18.1 & 39.0 & 10.8 & 50.2 & 37.2 & \cellcolor{Gray} \textbf{33.3} \\
            \hline
        \end{tabular} &
        \begin{tabular}[t]{|C{2cm}|ccccccc|}
            \hline
            \begin{tabular}{c} \textbf{DeiT-B} \\  \cite{touvron2021training} \end{tabular} & clp & inf & pnt & qdr & rel & skt & Avg. \\
            \hline
            clp  & -    & 24.3 & 49.6 & 15.8 & 65.3 & 52.1 & 41.4 \\
            inf  & 45.9 & -    & 45.9 & 6.7  & 61.4 & 39.5 & 39.9 \\
            pnt  & 53.2 & 23.8 & -    & 6.5  & 66.4 & 44.7 & 38.9 \\
            qdr  & 31.9 & 6.8  & 15.4 & -    & 23.4 & 20.6 & 19.6 \\
            rel  & 59.0 & 25.8 & 56.3 & 9.16 & -    & 44.8 & 39.0 \\
            skt  & 60.6 & 20.6 & 48.4 & 16.5 & 61.2 & -    & 41.5 \\
            Avg. & 50.1 & 20.3 & 43.1 & 10.9 & 55.5 & 40.3 & \cellcolor{Gray} \textbf{36.7} \\
            \hline
        \end{tabular} &
        \begin{tabular}[t]{|C{2cm}|ccccccc|}
            \hline
            \begin{tabular}{c} \textbf{SHOT-B} \\  \cite{SHOT} \end{tabular} & clp & inf & pnt & qdr & rel & skt & Avg. \\
            \hline
            clp  & -    & 27.0 & 49.7 & 16.5 & 65.4 & 53.2 & 46.1 \\
            inf  & 46.4 & -    & 45.9 & 7.4  & 60.6 & 40.1 & 40.1 \\
            pnt  & 54.6 & 25.7 & -    & 8.1  & 66.3 & 49.0 & 40.7 \\
            qdr  & 33.3 & 6.8  & 15.5 & -    & 23.8 & 24.0 & 20.7 \\
            rel  & 59.3 & 28.1 & 57.4 & 9.0  & -    & 47.3 & 40.2 \\
            skt  & 64.0 & 26.5 & 55.0 & 18.2 & 63.8 & -    & 45.5 \\
            Avg. & 51.5 & 26.6 & 44.7 & 11.8 & 56.0 & 42.7 & \cellcolor{Gray} \textbf{38.9} \\
            \hline
        \end{tabular} \\
        \begin{tabular}[t]{|C{2cm}|ccccccc|}
            \hline
            \begin{tabular}{c} \textbf{CDTrans$^*$} \\  \cite{xu2021cdtrans} \end{tabular} & clp & inf & pnt & qdr & rel & skt & Avg. \\
            \hline
            clp  & -    & 27.9 & 57.6 & 27.9 & 73.0 & 58.8 & 49.0 \\
            inf  & 58.6 & -    & 53.4 & 9.6  & 71.1 & 47.6 & 48.1 \\
            pnt  & 60.7 & 24.0 & -    & 13.0 & 69.8 & 49.6 & 43.4 \\
            qdr  & 2.9  & 0.4  & 0.3  & -    & 0.7  & 4.7  & 1.8  \\
            rel  & 49.3 & 18.7 & 47.8 & 9.4  & -    & 33.5 & 31.7 \\
            skt  & 66.8 & 23.7 & 54.6 & 27.5 & 68.0 & -    & 48.1 \\
            Avg. & 47.7 & 18.9 & 42.7 & 17.5 & 56.5 & 38.8 & \cellcolor{Gray} \textbf{37.0} \\
            \hline
        \end{tabular} &
        \begin{tabular}[t]{|C{2cm}|ccccccc|}
            \hline
            \begin{tabular}{c} \textbf{SSRT-B$^*$} \\  \cite{sun2022safe} \end{tabular} & clp & inf & pnt & qdr & rel & skt & Avg. \\
            \hline
            clp  & -    & 33.8 & 60.2 & 19.4 & 75.8 & 59.8 & 49.8 \\
            inf  & 55.5 & -    & 54.0 & 9.0  & 68.2 & 44.7 & 46.3 \\
            pnt  & 61.7 & 28.5 & -    & 8.4  & 71.4 & 55.2 & 45.0 \\
            qdr  & 42.5 & 8.8  & 24.2 & -    & 37.6 & 33.6 & 29.3 \\
            rel  & 69.9 & 37.1 & 66.0 & 10.1 & -    & 58.9 & 48.4 \\
            skt  & 70.6 & 32.8 & 62.2 & 21.7 & 73.2 & -    & 52.1 \\
            Avg. & 60.0 & 28.2 & 53.3 & 13.7 & 65.3 & 50.4 & \cellcolor{Gray} \textbf{45.2} \\
            \hline
        \end{tabular} &
        \begin{tabular}[t]{|C{2cm}|ccccccc|}
            \hline
            \begin{tabular}{c} \textbf{DSiT} \\  \textbf{(Ours)} \end{tabular} & clp & inf & pnt & qdr & rel & skt & Avg. \\
            \hline
            clp  & -    & 27.2 & 51.8 & 23.1 & 70.2 & 54.7 & 45.4 \\
            inf  & 52.3 & -    & 48.8 & 12.8 & 68.3 & 44.2 & 45.3 \\
            pnt  & 59.2 & 26.1 & -    & 14.5 & 71.5 & 51.4 & 44.5 \\
            qdr  & 38.1 & 8.3  & 21.2 & -    & 37.2 & 27.6 & 26.5 \\
            rel  & 60.4 & 28.0 & 57.8 & 13.1 & -    & 49.7 & 41.8 \\
            skt  & 66.3 & 27.5 & 56.0 & 24.4 & 70.2 & -    & 48.9 \\
            Avg. & 55.3 & 23.4 & 47.1 & 17.6 & 63.5 & 45.5 & \cellcolor{Gray} \textbf{42.1} \\
            \hline
        \end{tabular}
    \end{tabular}
    }
\end{table*}

%% file: suppl_tables/vendor_side.tex
\begin{table}[t]
\caption{Vendor-side Performance of Single-Source Domain Adaptation (SSDA) on Office-Home, DomainNet and VisDA.}
\setlength{\tabcolsep}{10pt}
\label{tab:vendor-side}
\resizebox{\columnwidth}{!}{
\begin{tabular}{cllll}
\toprule
\textbf{Training stage} &
  \multicolumn{1}{c}{\textbf{Method}} &
  \multicolumn{1}{c}{\textbf{\begin{tabular}[c]{@{}c@{}}Office-Home\\ (4 settings)\end{tabular}}} &
  \multicolumn{1}{c}{\textbf{VisDA}} &
  \multicolumn{1}{c}{\textbf{DomainNet}} \\ \midrule
\multirow{2}{*}{Vendor-side} & SHOT & 74.8        & 67.4        & 36.7        \\
                             & DSiT & 76.6 \textbf{\color{ForestGreen}{(+1.8)}} & 68.7 \textbf{\color{ForestGreen}{(+1.3)}} & 37.8 \textbf{\color{ForestGreen}{(+1.1)}} \\ \midrule
\multirow{2}{*}{Client-side} & SHOT & 79.0        & 85.9        & 38.3        \\
                             & DSiT & 81.1 \textbf{\color{ForestGreen}{(+2.1)}} & 87.5 \textbf{\color{ForestGreen}{(+1.6)}} & 42.1 \textbf{\color{ForestGreen}{(+3.8)}} \\ \bottomrule
\end{tabular}
}
\end{table}

%% file: suppl_tables/so_comp.tex
\begin{table}[t]
\caption{Single-Source Domain Adaptation (SSDA) on Office-Home (4 settings) for different vendor-side and client-side adaptation strategies.}
\label{tab:so_comp}
\resizebox{\columnwidth}{!}{
\begin{tabular}{@{}cccccccl@{}}
\toprule
\textbf{\#} & \textbf{Vendor-side} & \textbf{Client-side} & \textbf{Ar→Cl} & \textbf{Cl→Pr} & \textbf{Pr→Rw} & \textbf{Rw→Ar} & \multicolumn{1}{c}{\textbf{Avg}} \\ \midrule
1. & SHOT & SHOT & 67.1 & 83.4 & 85.3 & 80.4 & 79.1        \\
2. & SHOT & DSiT & 68.4 & 85.7 & 86.8 & 81.8 & 80.7 \textbf{\color{ForestGreen}{(+1.6)}} \\ \midrule
3. & DSiT & SHOT & 67.1 & 83.0 & 84.9 & 81.0 & 79.0        \\
4. & DSiT & DSiT & 69.2 & 86.8 & 86.6 & 82.4 & 81.3 \textbf{\color{ForestGreen}{(+2.3)}} \\ \bottomrule
\end{tabular}
}
\vspace{-4mm}
\end{table}

%% file: suppl_tables/vit_s.tex
\begin{table}[t]
    \centering
    \setlength{\tabcolsep}{3pt}
    \caption{Single-Source Domain Adaptation (SSDA) on Office-Home on ViT-S and DeiT-S Backbone. SF indicates \textit{source-free} adaptation. ``-S" denotes Small ViT Backbone.} 
    \label{tab:vit-s}
    \vspace{0.5mm}
    \resizebox{\columnwidth}{!}{
        \setlength{\extrarowheight}{1pt}
        \begin{tabular}{lccccccl}
            \toprule
            \multirow{2}{30pt}{\centering Method} & \multirow{2}{*}{\centering SF} &  &\multicolumn{5}{c}{\textbf{Office-Home}} \\
            \cmidrule{4-8}
            && & Ar$\shortrightarrow$Cl & Cl$\shortrightarrow$Pr & Pr$\shortrightarrow$Rw & Rw$\shortrightarrow$Ar & Avg. \\
            \midrule
            CDTrans-S \cite{xu2021cdtrans} & \xmark & & 60.7 & 75.6 & 84.4 & 77.0 & 74.4 \\
            \rowcolor{Gray} SHOT-S & \cmark & & \textbf{56.3} & 73.7 & 81.3 & 76.7 & 71.9 \\
            \rowcolor{Gray} \textbf{DSiT-S (\textit{Ours})} & \cmark & & {55.3} & \textbf{77.4} & \textbf{83.0} & \textbf{76.9} & \textbf{73.1} \textbf{\color{ForestGreen}{(+1.2)}}\\
            \bottomrule
            \rowcolor{Gray} SHOT-B & \cmark & & 69.07 & 85.31 & 88.13 & 83.89 & 81.6 \\
            \rowcolor{Gray} \textbf{DSiT-B (\textit{Ours})} & \cmark & & \textbf{71.84} & \textbf{87.18} & \textbf{88.11} & \textbf{83.4} & \textbf{82.6} \textbf{\color{ForestGreen}{(+1.0)}}\\
            \bottomrule
            \end{tabular} 
        }
\end{table}

%% file: tables/mtda_oh.tex
\begin{table}[t]
    \centering
    \setlength{\tabcolsep}{5.5pt}
    \setlength{\extrarowheight}{1pt}
    \caption{Multi-Target Domain Adaptation (MTDA) on Office-Home. SF indicates \textit{source-free} adaptation.
    ResNet-based methods (top) and Transformer-based methods (bottom).
    } 
    \label{tab:mtda_oh}
    \resizebox{\columnwidth}{!}{
        \begin{tabular}{lccccccl}
            \toprule
            \multirow{2}{30pt}{\centering Method} & \multirow{2}{*}{\centering SF} &  &\multicolumn{5}{c}{\textbf{Office-Home}} \\
            \cmidrule{4-8}
            && & Ar$\shortrightarrow$ & Cl$\shortrightarrow$ & Pr$\shortrightarrow$ & Rw$\shortrightarrow$ & Avg. \\
            \midrule
            MT-MTDA \cite{nguyen2021unsupervised} & \xmark &  & 64.6 & 66.4 & 59.2 & 67.1 & 64.3 \\
            CDAN+DCL \cite{CDAN} & \xmark & & 63.0 & 66.3 & 60.0 & 67.0 & 64.1 \\
            D-CGCT \cite{D-CGCT} & \xmark & & 70.5 & 71.6 & 66.0 & 71.2 & 69.8 \\
            \midrule
            D-CGCT-B \cite{D-CGCT} & \xmark & & 77.0 & 78.5 & 77.9 & 80.9 & 78.6 \\
            \rowcolor{Gray} SHOT-B* & \cmark & & 75.4 & 79.3 & 73.6 & 77.1 & 76.4 \\
            \rowcolor{Gray} \textbf{DSiT-B (\textit{Ours})} & \cmark & & \textbf{77.3} & \textbf{83.4} & \textbf{75.6} & \textbf{76.8} & \textbf{78.3} \textbf{\color{ForestGreen}{(+1.9)}}\\
            \bottomrule
            \end{tabular} 
        }
        \vspace{-2mm}
\end{table}

%% file: suppl_tables/train_time.tex
\begin{table}[h]
    \centering
    \setlength{\tabcolsep}{3.5pt}
    \caption{Training time comparison of our approach DSiT vs SHOT on Office-Home (Rw$\shortrightarrow$Ar)}
    \label{tab:training_time}
    \resizebox{\columnwidth}{!}{
        \setlength{\extrarowheight}{1pt}
        \begin{tabular}{lccccccc}
            \toprule
            \multirow{3}{45pt}{\centering \textbf{Method}} & \multicolumn{5}{c}{\textbf{Training time (in min)}} & \multirow{2}{20pt}{\begin{tabular}{c}\textbf{Inf.}\\\textbf{time}\\\textbf{(ms)}\end{tabular}} & \multirow{4}{20pt}{\textbf{Acc.}} \\
            \cmidrule{2-6}
            & \begin{tabular}{c}Src. \\ train\end{tabular} & \begin{tabular}{c}Src. \\ DST\end{tabular} & \begin{tabular}{c}Tgt.\\adapt\end{tabular} & \begin{tabular}{c}Tgt.\\DST\end{tabular} & \begin{tabular}{c}Total\\time\end{tabular}&&\\
            \midrule

            SHOT-B    & 12 & - & 17 & - & 29 & 3.6 & 80.4 \\
            \emph{Ours} & 12 & 109 & - & 258 & 270 & 3.6 & \textbf{82.4} \\
            \bottomrule
            \end{tabular} 
        }
        \vspace{-3mm}
\end{table}

%% file: suppl_tables/a_dist.tex
\begin{table}[ht]
    \centering
    \setlength{\tabcolsep}{12pt}
    \caption{Analysis for $\mathcal{A}$-distance of three augmentations on 4 settings of Office-Home (SSDA).}
    \setlength{\extrarowheight}{1pt}
    \label{tab:a-dist augs}
    \resizebox{\columnwidth}{!}{
        \begin{tabular}{lcccc}
            \toprule
            \textbf{Aug.} & \textbf{Ar$\shortrightarrow$Cl} & \textbf{Cl$\shortrightarrow$Pr} & \textbf{Pr$\shortrightarrow$Rw} & \textbf{Rw$\shortrightarrow$Ar} \\
            \midrule
            FDA  & 0.857 & 0.730 & 0.829 & 0.286 \\
            \textbf{Original}	& \textbf{1.049} & \textbf{0.852} & \textbf{0.834} & \textbf{0.504} \\
            AdaIN       & 1.072 & 0.648 & 0.842 & 0.136 \\
            \bottomrule
        \end{tabular} 
        }
\end{table}

%% file: suppl_tables/task_epochs.tex
\begin{table}[ht]
    \centering
    \setlength{\tabcolsep}{5.3pt}
    \caption{Sensitivity Analysis on Single-Source Domain Adaptation (SSDA) on Office-Home. (4 settings)}
    \label{tab:sensitivity_analysis}
    \resizebox{\columnwidth}{!}{
        \begin{tabular}{cccccc}
            \toprule
            \textbf{Epochs} & \textbf{Ar $\shortrightarrow$ Cl} & \textbf{Cl $\shortrightarrow$ Pr} & \textbf{Pr $\shortrightarrow$ Rw} & \textbf{Rw $\shortrightarrow$ Ar} & \textbf{Avg.} \\
            \midrule
            1 & 64.1 & 79.8 & 84.7 & 79.6 & 77.0\\
            2 & 69.2 & 86.1 & 86.6 & 82.4 & 81.1\\
            3 & 69.6 & 86.7 & 87.3 & 82.4 & 81.5\\
            5 & 69.5 & 86.3 & 87.1 & 82.5 & 81.3\\
            \bottomrule
        \end{tabular} 
        }
\end{table}

%% file: suppl_tables/significance.tex
\begin{table}[ht]
    \centering
    \setlength{\tabcolsep}{9pt}
    \caption{Significance experiments of DSiT-B (Ours) on Single-Source DA (SSDA) on Office-Home (4 settings).  }
    \label{tab:mean_std}
    \resizebox{\columnwidth}{!}{
        \begin{tabular}{ccccc}
            \toprule
             \textbf{Ar $\shortrightarrow$ Cl} & \textbf{Cl $\shortrightarrow$ Pr} & \textbf{Pr $\shortrightarrow$ Rw} & \textbf{Rw $\shortrightarrow$ Ar} & \textbf{Avg.} \\
            \midrule
            69.2 \small{$\pm$0.1} & 86.1 \small{$\pm$0.3} & 86.6 \small{$\pm$0.3} & 82.4 \small{$\pm$0.6} & 81.1 {$\pm$0.1} \\
            \bottomrule
        \end{tabular} 
        }
        \vspace{-2mm}
\end{table}

%% file: suppl_tables/loss_abl.tex
\begin{table}[h]
    \centering
    \setlength{\tabcolsep}{5pt}
    \caption{Ablation study for the three components of the client-side adaptation on 4 settings of Office-Home. \textit{PL} indicates pseudo-labeling} 
    \label{tab:ablation_loss}
    \resizebox{\columnwidth}{!}{
        \begin{tabular}{lccccc}
        \toprule
        \textbf{Method} & \textbf{Ar $\shortrightarrow$ Cl} & \textbf{Cl $\shortrightarrow$ Pr} & \textbf{Pr $\shortrightarrow$ Rw} & \textbf{Rw $\shortrightarrow$ Ar} & \textbf{Avg.} \\
        \midrule
        Source Baseline & 62.5 & 79.4 & 84.3 & 79.2 & 76.4 \\
        $\mathcal{L}_{im}$ & 62.1 & 79.7 & 80.1 & 73.9 & 74.0 \\
        $\mathcal{L}_{im} + \mathcal{L}_{div}$ & 68.2 & 86.0 & 86.6 & 81.3 & 80.5 \\
        $\mathcal{L}_{im} + \mathcal{L}_{div} + \textit{PL}$ & 69.2 & 86.1 & 86.6 & 82.4 & 81.1 \\
        \bottomrule
        \end{tabular}
        }
        \vspace{-2mm}
\end{table}